\documentclass[10pt,twocolumn,letterpaper]{article}
\usepackage[pagenumbers]{wacv} 
\usepackage{amsmath,amssymb}
\usepackage{booktabs}
\usepackage{multirow}
\usepackage{array}
\usepackage{tabularx}
\usepackage{makecell}
\usepackage{placeins}
\usepackage{xcolor}
\usepackage{microtype}

\definecolor{wacvblue}{rgb}{0.21,0.49,0.74}
\usepackage[pagebackref,breaklinks,colorlinks,allcolors=wacvblue]{hyperref}

\title{PhysioAI: Clinical Knowledge-Guided Semantic Supervision for Skeleton-Based Physiotherapy Action Recognition}

\author{
Jie Cao$^{1}$ \quad
Euijoon Ahn$^{2}$ \quad
Anwar Hassan$^{3}$ \quad
Jinman Kim$^{1}$\\[4pt]
{\small $^{1}$The University of Sydney, Australia}\\
{\small $^{2}$James Cook University, Australia}\\
{\small $^{3}$Nepean Hospital, Australia}
}

\begin{document}
\maketitle
\begin{abstract}
Skeleton-based action recognition can support automated tracking of physiotherapy exercises, particularly in remote rehabilitation settings where continuous in-person supervision is impractical. However, most existing methods are developed for large-scale daily-action benchmarks rather than rehabilitation scenarios. Public rehabilitation exercise datasets are typically small, with only subtle kinematic differences between exercise classes. For participants with motor impairments, exercise execution may also deviate from standard movement patterns in amplitude, speed, and coordination, increasing intra-class variability and making reliable recognition more difficult for skeleton-based models. We propose \emph{PhysioAI}, a clinical knowledge-guided semantic supervision framework that injects structured physiotherapy knowledge into skeleton representation learning. PhysioAI combines graph-based spatiotemporal modelling of human movement with training-time semantic anchors derived from a structured Clinical Knowledge Dictionary (CKD). The CKD descriptions are encoded using a frozen Contrastive Language–Image Pre-training (CLIP) model and projected into an anchor space, where they provide class-specific semantic targets for skeleton representation learning. The resulting CKD-derived anchors are used only during skeleton-model training; inference requires only skeleton inputs. Under subject-disjoint evaluation, PhysioAI achieves $99.03{\pm}1.34\%$ on KiMoRe Overall, $94.64{\pm}7.36\%$ on the Hard-67 stress test, and $87.44{\pm}7.69\%$ on UI-PRMD Overall. These results exceed the strongest comparator for each endpoint by $0.27$, $2.87$, and $1.33$ percentage points (pp), respectively. These findings demonstrate that structured clinical knowledge can serve as an effective source of training-time supervision for physiotherapy action recognition.
\end{abstract}
\vspace{-12pt}
\section{Introduction}
\label{sec:intro}

Telerehabilitation has emerged as an effective approach for delivering physiotherapy remotely, using video-based communication to overcome geographic, mobility, scheduling, and resource barriers to supervised exercise \cite{lee_telerehabilitation_2024}. However, continuous manual observation of exercise performance becomes impractical as the number of remotely supervised sessions increases, while patients frequently perform prescribed exercises independently between therapist interactions. Automated recognition of physiotherapy exercises from video-derived human motion can support therapist oversight, patient self-management, adherence assessment, and downstream rehabilitation applications in such settings.

\begin{figure*}[!t]
    \centering\includegraphics[width=0.74\textwidth]{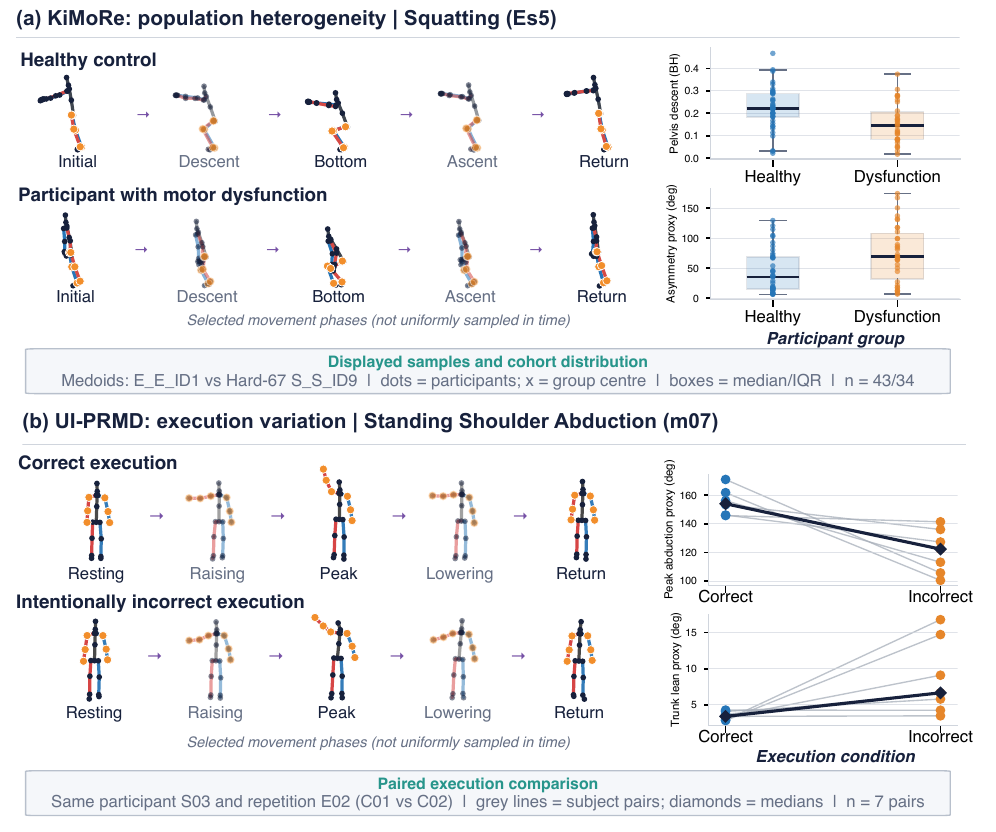}
    \caption{\textbf{Population and execution variability in rehabilitation movements.} The skeleton strips show five selected movement phases in temporal order. (a) KiMoRe Squatting (Es5): healthy-control and motor-dysfunction motion examples, together with skeleton-derived pelvis-descent and asymmetry distributions for participants with available Es5 sequences ($n=43$ healthy controls and $n=34$ participants with motor dysfunction). In the right-hand plots, x-axis denotes participant group; dots denote participants, and boxes show the median and interquartile range. The displayed motor-dysfunction sequence is a deterministically selected medoid from the fixed Hard-67 samples misclassified by CTR-GCN, whereas the distributions include all eligible participants. (b) UI-PRMD Standing Shoulder Abduction (m07): a same-participant, same-repetition pair of correct and intentionally incorrect executions, with paired skeleton-derived abduction and trunk-lean proxies for seven participants with complete condition pairs. UI-PRMD was collected from healthy participants who intentionally performed both execution conditions; the incorrect trials are recorded movements rather than synthetically generated sequences. In the right-hand plots, the x-axis denotes execution condition (correct or incorrect); lines connect participant-level paired values, and diamonds denote medians. These quantities are observable skeleton-derived proxies rather than clinical movement-quality scores.}
    \label{fig:execution_variability}
\end{figure*}

Skeleton-based action recognition represents the human body as a graph of joints and seeks to learn discriminative spatiotemporal motion patterns directly from skeletal trajectories. Such representations are particularly attractive for physiotherapy monitoring because they provide a compact, motion-centric description while suppressing much of the appearance information present in RGB video. Physiotherapy exercises are often distinguished only by subtle kinematic characteristics, including the body regions involved, movement planes, joint actions, and movement phases. At the same time, the same exercise may be performed using diverse movement strategies due to differences in physical capability, compensatory behaviours, or motor impairments. Consequently, rehabilitation datasets exhibit both fine-grained inter-class ambiguity and substantial intra-class execution variability, creating challenging conditions for reliable action recognition. Furthermore, available rehabilitation resources such as KiMoRe~\cite{capecci_2019_the} and UI-PRMD~\cite{vakanski_data_2018} remain small relative to general action-recognition benchmarks, further limiting the data available for training. Figure~\ref{fig:execution_variability} illustrates complementary sources of variation across participant populations and execution conditions.

These conditions highlight the need for representations that capture how exercises differ beyond their class identities. Graph convolutional networks (GCNs) model the body as a spatiotemporal graph and form a strong basis for skeleton action recognition~\cite{yan_2018_spatial,shi_2019_twostream,chen_2021_channelwise,liu_disentangling_2020}, but their supervision is typically dominated by categorical labels. Recent language-guided methods show that textual descriptions can enrich skeleton representation learning~\cite{xiang_generative_2023,zhu_part-aware_2024,chen_linguistic-driven_2024,xu_language_2025}, but they primarily target general-purpose action recognition and do not resolve how semantic supervision should be constructed and controlled in the rehabilitation setting. Directly transferring these strategies raises several design issues. Generic action descriptions may omit the observable kinematic cues that distinguish confusable physiotherapy exercises; limited per-class data make it difficult to jointly learn stable discriminative skeleton representations and reliable cross-modal mappings. This motivates a practical question: how can rehabilitation-specific semantics be constructed and incorporated into skeleton representation learning while accounting for both fine-grained class distinctions and limited per-class supervision?

This paper makes three contributions. First, we present PhysioAI, a clinical knowledge-guided semantic supervision framework for skeleton-based physiotherapy exercise recognition with skeleton-only deployment. Second, we introduce clinical knowledge-guided semantic supervision for skeleton representation learning through structured Clinical Knowledge Dictionary (CKD) anchors and uncertainty-weighted semantic alignment. Third, we evaluate PhysioAI under subject-disjoint protocols on two public rehabilitation datasets and on Hard-67, a stress test derived from the training dynamics of a Spatial Temporal Graph Convolutional Network (ST-GCN) baseline~\cite{yan_2018_spatial}. Among the evaluated methods, PhysioAI achieves the highest observed values across the reported endpoints on both datasets and Hard-67.

\section{Related Work}
\label{sec:related}
\subsection{Physiotherapy Exercise Analysis}

Public datasets such as KiMoRe and UI-PRMD support two related tasks: recognising which physiotherapy exercise is performed and assessing how well it is performed~\cite{capecci_2019_the,vakanski_data_2018}. This work focuses on exercise recognition.

Non-graph methods typically use Convolutional Neural Networks (CNNs), Recurrent Neural Networks (RNNs), or hybrid temporal architectures. Zaher et al.~\cite{zaher_unlocking_2025}, for example, evaluated CNN, Long Short-Term Memory (LSTM), Bidirectional LSTM (Bi-LSTM), and CNN--LSTM models on KiMoRe and UI-PRMD, demonstrating the value of temporal modelling for physiotherapy exercise recognition. These architectures model motion over time but do not explicitly represent the body as a graph of interacting joints.

Graph-based approaches directly model skeletal connectivity and inter-joint motion. Ensemble-based Graph Convolutional Network (EGCN)~\cite{yu_egcn_2022} explores data, feature, decision, and model-level fusion of skeleton position and orientation streams for physiotherapy exercise assessment. However, most physiotherapy studies focus on quality assessment or score prediction rather than action recognition, leaving difficult cases arising from variations in exercise execution less explored. PhysioAI addresses this gap by focusing on physiotherapy exercise recognition under such variability.

\subsection{Skeleton-Based Action Recognition}
GCNs have become a dominant approach for action recognition from skeletal motion data due to their ability to model both spatial joint relationships and temporal dynamics. ST-GCN established graph convolution over spatial joints and temporal connections~\cite{yan_2018_spatial}. Subsequent models introduced adaptive graphs~\cite{shi_2019_twostream}, multi-scale spatiotemporal aggregation~\cite{liu_disentangling_2020}, and part-aware residual modelling~\cite{song_stronger_2020}. Channel-wise Topology Refinement Graph Convolution (CTR-GCN) learns a shared graph topology and refines it using channel-specific joint correlations, allowing different feature channels to model distinct relational patterns~\cite{chen_2021_channelwise}. This flexible topology modelling provides the backbone used in PhysioAI. More recently, Prototypical Graph Convolutional Network (ProtoGCN) incorporated learnable prototypes to capture local motion patterns and action-specific dynamics~\cite{liu_revealing_2025}. 

Despite their strong performance, these models were primarily developed for large-scale daily-action benchmarks such as NTU RGB+D~\cite{shahroudy_2016_ntu}. Their applicability to rehabilitation recognition has not been systematically established under the combined conditions of limited training data, fine-grained exercise distinctions, and participant-dependent execution variability. Rather than introducing another graph operator, PhysioAI uses CTR-GCN as its skeleton encoder and focuses on incorporating structured rehabilitation semantics into representation learning.

\subsection{Language-guided Skeleton Learning}
Language-guided methods provide skeleton models with richer descriptions of action structure than class labels alone. Generative Action-description Prompts (GAP) generates global and body-part action descriptions and aligns their text embeddings with skeleton features during supervised training~\cite{xiang_generative_2023}. Part-aware Unified Representation of Language and Skeleton (PURLS) learns global and local language--skeleton alignment for zero-shot recognition, where prediction relies on semantic prototypes of unseen classes~\cite{zhu_part-aware_2024}. Linguistic-Driven Partial Semantic Relevance Learning (LPSR) incorporates large language model (LLM)-generated global and local movement descriptions through semantic consistency and cross-modal feature interaction~\cite{chen_linguistic-driven_2024}, whereas Language Knowledge-Assisted Graph Convolutional Network (LA-GCN) converts language knowledge into graph-based structural priors~\cite{xu_language_2025}. These methods demonstrate that language can guide skeleton representation learning. Unlike language assets designed for general action recognition, PhysioAI derives its semantic supervision from structured physiotherapy exercise definitions.

\section{Method}
\label{sec:method}

\begin{figure*}[!t]
    \centering
    \includegraphics[width=0.84\textwidth]
    {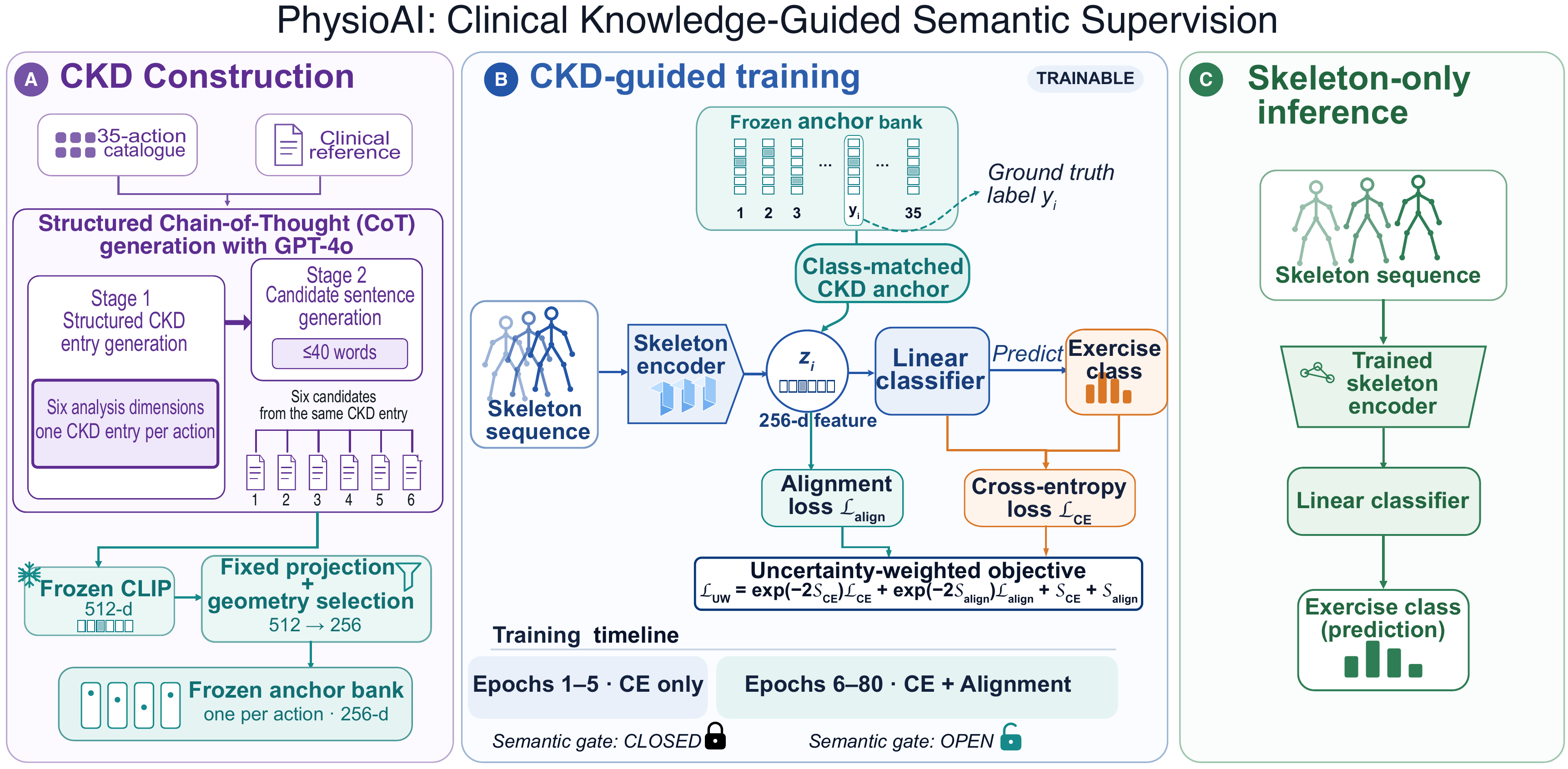}
    \caption{\textbf{Overview of our PhysioAI framework.}
    A: CKD descriptions are prepared before model training, encoded by Contrastive Language–Image Pre-training (CLIP), and converted into fixed class-specific semantic anchors. B: During training, the skeleton encoder is optimised using exercise classification and class-matched semantic alignment. Alignment is activated at epoch~6 and balanced with classification through uncertainty weighting. C: At inference, only the trained skeleton encoder and classifier are retained.}
    \label{fig:architecture}
\end{figure*}

\subsection{Framework Overview}
\label{sec:framework}

An overview of PhysioAI is shown in Figure~\ref{fig:architecture}. First, a CKD is constructed before skeleton-model training, and one discriminative sentence per class is encoded by a frozen CLIP text encoder~\cite{radford_2021_learning} and projected into skeleton-feature space to form fixed semantic anchors. Second, a CTR-GCN joint-stream encoder~\cite{chen_2021_channelwise} is trained using classification and class-matched semantic alignment, with alignment activated at epoch~6 and balanced through uncertainty weighting. Third, the semantic components are discarded, leaving only the skeleton encoder and classifier for inference.

\subsection{Clinical Knowledge Dictionary}
\label{sec:ckd}
\paragraph{Knowledge sources and exercise catalogue.} The CKD schema and source hints draw on three groups of public sources: foundational kinematics and kinesiology references~\cite{aaos_1965_joint,norkin_2016_measurement,neumann_2017_kinesiology}, rehabilitation practice guidelines~\cite{winstein_2016_guidelines,george_2020_clinical,osborne_2022_physical,nice2017parkinson}, and gait-analysis references~\cite{winter_2009_biomechanics,perry_2010_gait}. It contains 35 rehabilitation and functional-action entries across five domains: functional movement screening, lower-limb exercise, upper-limb exercise, balance and postural control, and gait or locomotion. Each entry records an action identifier and structured movement descriptors, including movement taxonomy, primary body region, movement plane, joint action, support pattern, laterality, movement phases, and cues that distinguish the action from similar classes.

\paragraph{Constrained CKD generation.} CKD entries are generated before skeleton-model training using a six-step structured Chain-of-Thought (CoT) prompting procedure~\cite{wei_chain--thought_2022}. In the first stage, GPT-4o with temperature~0.2 applies a fixed CKD schema to a frozen 35-action catalogue and associated source hints. The schema organises movement taxonomy, anatomy, and body-part kinematics, confusion discriminators, population-specific execution variations, skeleton-joint correspondence, and phase-level movement landmarks. In the second stage, GPT-4o with temperature~0.5 reads the resulting structured fields and generates six candidate discriminative sentences of at most 40 words for each action. The generation prompt requires candidates to prioritise class-stable cues observable from skeletons, including the primary moving region, movement plane, joint action, support pattern, laterality, and key movement phases, while excluding pathology wording and observer-facing instructions.

\paragraph{Sentence selection and CLIP encoding.} Before skeleton-model training, all candidate sentences are encoded using a frozen CLIP ViT-B/32 text encoder~\cite{radford_2021_learning}. The resulting normalised 512-dimensional embeddings are mapped to 256 dimensions as described in Section~\ref{sec:semantic_supervision}. A fixed text-geometry objective is then used to select one sentence for each action. The objective considers the mean similarity across all anchor pairs, the mean similarity of 15 predefined confusable action pairs, mean nearest-neighbour similarity, maximum pairwise similarity, mean similarity within predefined action families, and the number of anchor pairs whose cosine similarity exceeds 0.90. A lower objective value indicates a more separable projected text-anchor bank.

Sentence selection is performed by greedy coordinate descent in the fixed 35-action catalogue order. A candidate replaces the current sentence only when it strictly reduces the combined text-geometry and sentence-format score. The sentence-format penalty discourages observer-directed openings, missing declarative action labels, and incomplete sentence termination. The complete generation prompts, 35-action catalogue, candidate sentences, objective coefficients, sentence-format penalty, confusable action pairs, action-family assignments, and selection algorithm are provided in the supplementary material.

For example, the structured CKD entry for \emph{Deep Squat} describes bilateral hip, knee, and ankle flexion in the sagittal plane, an overhead arm position, distinguishing cues, relevant joints, and four movement phases. These fields are used to generate six candidate sentences, whose projected CLIP embeddings are evaluated jointly with the other action anchors. The procedure retains \emph{``Deep squat action: Begin with sagittal plane flexion of hips, knees, and ankles, maintaining arms overhead for balance.''} The complete candidates are provided in the supplementary material.

For CKD class $c$, the selected sentence is represented by a normalised 512-dimensional CLIP embedding $a_c$. Before training on a particular dataset, a fixed class-to-CKD mapping connects each dataset label index to its semantic anchor. KiMoRe therefore uses five CKD embeddings, while UI-PRMD uses ten.

\subsection{CKD-Guided Semantic Supervision}
\label{sec:semantic_supervision}

\paragraph{Fixed Gaussian projection.} Section~\ref{sec:ckd} provides a normalised 512-dimensional CLIP embedding $a_c$ for each of the $K$ exercise classes, where $c\in\{1,\ldots,K\}$. Because the skeleton encoder produces a $d$-dimensional pre-classifier feature with $d=256$, we use a fixed Gaussian random projection applied during sentence selection to map the selected text embeddings into the skeleton-feature space~\cite{bingham_random_2001}. A Gaussian matrix $P\in\mathbb{R}^{512\times d}$ is column-normalised and used to construct the class anchor
\begin{equation}
    b_c =
    \frac{a_cP}
         {\lVert a_cP\rVert_2},
    \qquad
    b_c\in\mathbb{R}^{d}.
    \label{eq:projection}
\end{equation}
Here, $b_c$ is a fixed class-specific semantic direction used by the cosine-alignment objective. The projection and resulting anchors remain fixed throughout training and introduce no trainable parameters. The projection serves only as a dimension adapter; the relationship between skeleton features and CKD anchors is learned through semantic alignment.

\paragraph{Class-matched semantic alignment.} For a training batch $\mathcal{B}$, sample $i$ contains a skeleton sequence $X_i$ and its labelled exercise class $y_i\in\{1,\ldots,K\}$. The skeleton encoder $f_{\theta}$, with trainable parameters $\theta$, maps $X_i$ to the pre-classifier feature $z_i=f_{\theta}(X_i)\in\mathbb{R}^{d}$. We align this feature with the CKD anchor associated with its labelled class:
\begin{equation}
    \mathcal{L}_{\mathrm{align}}
    =
    \frac{1}{|\mathcal{B}|}
    \sum_{i\in\mathcal{B}}
    \left(
    1-
    \frac{z_i^{\mathsf{T}}b_{y_i}}
         {\lVert z_i\rVert_2}
    \right),
    \label{eq:alignment}
\end{equation}
where $|\mathcal{B}|$ is the number of samples in the training batch and $b_{y_i}$ is the anchor corresponding to class $y_i$. Because $b_{y_i}$ is already $\ell_2$-normalised, the inner-product term in Eq.~\ref{eq:alignment} is the cosine similarity between the skeleton feature and its class anchor. Minimising $\mathcal{L}_{\mathrm{align}}$ therefore guides the skeleton representation towards the fixed semantic direction associated with its labelled exercise class. The anchors remain fixed, while the skeleton encoder learns features that align with them. Section~\ref{sec:delayed_uw} describes how this alignment loss is combined with the classification objective.

\subsection{Delayed Uncertainty-Weighted Optimisation}
\label{sec:delayed_uw}

The skeleton feature $z_i$ is passed to a linear classifier $g_{\phi}$ with trainable parameters $\phi$, producing logits $o_i=g_{\phi}(z_i)\in\mathbb{R}^{K}$ for the $K$ exercise classes. The standard cross-entropy (CE) loss computed from the logits and labels in training batch $\mathcal{B}$ is written as $\mathcal{L}_{\mathrm{CE}}$, while $\mathcal{L}_{\mathrm{align}}$ is the semantic-alignment loss defined in Eq.~\ref{eq:alignment}.

To balance the classification and semantic-alignment objectives without manually selecting a fixed loss coefficient, PhysioAI adopts a log-scale parameterisation adapted from homoscedastic uncertainty weighting (UW)~\cite{kendall_multi-task_2018}. Two learnable scalar parameters, $s_{\mathrm{ce}}$ and $s_{\mathrm{align}}$, control the contributions of the classification and alignment objectives, respectively. These are global training parameters rather than sample-specific uncertainty estimates. Both are initialised to zero, giving an initial effective weight of one when the corresponding objective is active:
\begin{align}
    U_{\mathrm{CE}}
    &=
    e^{-2s_{\mathrm{ce}}}
    \mathcal{L}_{\mathrm{CE}}
    +s_{\mathrm{ce}},
    \\
    U_{\mathrm{align}}
    &=
    e^{-2s_{\mathrm{align}}}
    \mathcal{L}_{\mathrm{align}}
    +s_{\mathrm{align}}.
\end{align}

Here, $U_{\mathrm{CE}}$ and $U_{\mathrm{align}}$ are the uncertainty-weighted classification and alignment objectives. The exponential parameterisation keeps each effective weight positive: increasing $s$ reduces the contribution of its corresponding loss, whereas decreasing $s$ increases it. The additive $s$ terms prevent the model from suppressing an objective simply by increasing its log-scale parameter without bound.

Jointly optimised objectives can produce conflicting gradients~\cite{yu_gradient_2020}. Based on this consideration, PhysioAI uses a fixed five-epoch classification-only burn-in aligned with the end of the five-epoch learning-rate warm-up in our CTR-GCN training recipe. Semantic alignment is activated at epoch~6. We treat this onset as a fixed scheduling choice, rather than a universally optimal value, and compare it with immediate activation in Section~\ref{sec:ablations}. For training epoch $t\in\{1,\ldots,80\}$, the complete objective is
\begin{equation}
\mathcal{L}(t)=
\begin{cases}
U_{\mathrm{CE}}, & 1\leq t\leq5,\\[2pt]
U_{\mathrm{CE}}+U_{\mathrm{align}}, & 6\leq t\leq80.
\end{cases}
\label{eq:schedule}
\end{equation}

During epochs~1--5, neither the semantic-alignment loss nor its associated $s_{\mathrm{align}}$ term contributes to optimisation. From epoch~6 onward, the complete alignment objective is included, and its relative contribution is controlled by the learned uncertainty weight. The effect of this schedule is evaluated in the ablation study.

\section{Experiments}
\label{sec:experiments}

\subsection{Datasets}

\paragraph{KiMoRe.} KiMoRe was recorded with a Kinect v2 sensor and includes 78 participants: 44 healthy participants and 34 participants with motor dysfunctions~\cite{capecci_2019_the}. KiMoRe was used for the primary recognition evaluation and the Hard-67 stress test because it captures exercise-class distinctions and movement variation across healthy participants and participants with motor impairments.

\paragraph{UI-PRMD.} UI-PRMD contains ten rehabilitation movements performed by ten healthy participants and recorded using Kinect and Vicon motion-capture systems~\cite{vakanski_data_2018}. Participants performed the demonstrated movements and intentionally non-optimal variants. Correct and deliberately incorrect executions of each exercise share one action label, as our task is exercise recognition rather than movement-quality assessment. UI-PRMD complements KiMoRe with a larger action vocabulary and controlled within-class execution variation.

\subsection{Implementation Details}
Most experiments used Python~3.10.19, PyTorch~2.5.1, CUDA~12.1, and cuDNN~9.1; ProtoGCN used Python~3.7.11, PyTorch~1.11.0, CUDA~11.3, and cuDNN~8.2. All ran on a single NVIDIA GeForce RTX~3090 GPU with 24~GB memory. All skeleton backbones and classifiers were trained from scratch without pretrained action-recognition checkpoints, using backbone-specific temporal lengths. Unless otherwise stated, training used 80 epochs, batch size 32, and Nesterov SGD with momentum~0.9 and weight decay $5\times10^{-4}$. The learning rate was linearly warmed to~0.1 over five epochs and decayed tenfold at epochs 35, 55, and 75. Complete per-method configurations and reproducibility settings are provided in the supplementary material.

\subsection{Hard-67 Stress-Test Protocol}
\label{sec:hard67}
Hard-67 is a fixed subset of KiMoRe constructed from the training dynamics of plain ST-GCN~\cite{yan_2018_spatial} using Dataset Cartography statistics~\cite{swayamdipta_dataset_2020}. ST-GCN is used only to define the subset; PhysioAI and the other comparison methods do not contribute to its construction. Each KiMoRe sample appears in the training partition of four folds and contributes one non-augmented true-class probability per epoch, giving 320 observations across four folds and 80 epochs. Confidence is calculated as the mean of these observations and variability as their standard deviation. The selection rule first takes samples at or below the 33rd percentile of confidence and then retains those at or below the median variability of this low-confidence group, yielding 67 from 400 samples.

Each Hard-67 sample is evaluated in its subject's test fold using the earliest checkpoint attaining the highest held-out Overall accuracy across epochs. We report the mean and sample standard deviation (SD) across folds, together with the pooled correct/67 count.

\subsection{Evaluation and Comparison}
\label{sec:results}

Overall accuracy was the primary evaluation metric. Macro-F1 (mF1) and balanced accuracy (BA) were also reported to account for class-level performance. Metrics were calculated separately for each subject-disjoint fold and summarised using the unweighted mean and SD. For class-level analysis, the held-out predictions from all five folds were combined so that each sequence contributes only once.

We compared PhysioAI with representative skeleton-based benchmarks including ST-GCN~\cite{yan_2018_spatial}, attention-enhanced adaptive graph convolutional network (AAGCN)~\cite{shi_multistream_2020}, Multi-Scale 3D Graph Convolutional Network (MS-G3D)~\cite{liu_disentangling_2020}, Part-wise Attention Residual GCN (PA-ResGCN)~\cite{song_stronger_2020}, CTR-GCN~\cite{chen_2021_channelwise}, BlockGCN~\cite{zhou_blockgcn_2024}, Skeletal-Temporal Transformer (SkateFormer)~\cite{do_skateformer_2024}, ProtoGCN~\cite{liu_revealing_2025}, and the supervised language-guided GAP~\cite{xiang_generative_2023}. All methods used the same subject partitions, evaluation metrics, and checkpoint-selection rule. GAP implementation followed the released method and prompt structure, with newly generated rehabilitation-action descriptions frozen before training. Our implementations were based on the released code or model specifications of the corresponding methods, and the comparison code will be released.

\begin{table*}[!t]
\centering
\caption{Comparison under subject-disjoint five-fold evaluation. Values are mean$\pm$SD (\%); Hard pool reports the number of correctly classified samples out of 67. Macro-F1 (mF1) and balanced accuracy (BA) were recomputed from the audited out-of-fold predictions. Bold and underline mark the best and second-best endpoint values, respectively; ties share formatting. Formatting does not indicate statistical significance.}
\label{tab:main}
\scriptsize
\setlength{\tabcolsep}{2.4pt}
\begin{tabular}{lcccccccc}
\toprule
& \multicolumn{3}{c}{KiMoRe} & \multicolumn{2}{c}{Hard-67} & \multicolumn{3}{c}{UI-PRMD} \\
\cmidrule(lr){2-4}\cmidrule(lr){5-6}\cmidrule(lr){7-9}
Method & Overall & mF1 & BA & Fold mean & Pool & Overall & mF1 & BA \\
\midrule
\multicolumn{9}{l}{\emph{Skeleton-only supervision}} \\
ST-GCN~\cite{yan_2018_spatial} & $95.86{\pm}2.18$ & $95.91{\pm}2.17$ & $95.98{\pm}2.15$ & $75.81{\pm}12.82$ & 52/67 & $84.50{\pm}2.66$ & $82.02{\pm}5.15$ & $83.15{\pm}3.93$ \\
AAGCN (joint)~\cite{shi_multistream_2020} & $75.27{\pm}11.01$ & $73.68{\pm}12.78$ & $75.09{\pm}11.03$ & $48.92{\pm}22.79$ & 36/67 & $79.64{\pm}7.82$ & $78.35{\pm}9.19$ & $79.81{\pm}8.02$ \\
MS-G3D~\cite{liu_disentangling_2020} & $\underline{98.76{\pm}1.81}$ & $\underline{98.78{\pm}1.78}$ & $\underline{98.83{\pm}1.69}$ & $90.18{\pm}18.66$ & \underline{63/67} & $85.01{\pm}5.82$ & $84.12{\pm}6.52$ & $84.88{\pm}5.78$ \\
PA-ResGCN~\cite{song_stronger_2020} & $96.72{\pm}1.23$ & $96.77{\pm}1.25$ & $96.82{\pm}1.12$ & $79.15{\pm}13.09$ & 55/67 & $79.72{\pm}5.05$ & $77.95{\pm}5.75$ & $78.58{\pm}4.60$ \\
CTR-GCN~\cite{chen_2021_channelwise} & $97.68{\pm}2.56$ & $97.67{\pm}2.59$ & $97.74{\pm}2.50$ & $84.82{\pm}24.02$ & 60/67 & $85.85{\pm}8.94$ & $\underline{85.40{\pm}9.51}$ & $86.26{\pm}8.51$ \\
BlockGCN~\cite{zhou_blockgcn_2024} & $97.28{\pm}1.92$ & $97.26{\pm}1.90$ & $97.31{\pm}1.81$ & $84.60{\pm}16.46$ & 59/67 & $66.30{\pm}14.52$ & $62.85{\pm}14.84$ & $64.98{\pm}14.51$ \\
SkateFormer~\cite{do_skateformer_2024} & $44.28{\pm}18.53$ & $37.79{\pm}21.02$ & $44.44{\pm}18.08$ & $27.68{\pm}16.28$ & 20/67 & $51.99{\pm}20.43$ & $47.07{\pm}23.98$ & $51.83{\pm}20.76$ \\
ProtoGCN~\cite{liu_revealing_2025} & $98.16{\pm}1.56$ & $98.19{\pm}1.53$ & $98.20{\pm}1.47$ & $90.18{\pm}18.66$ & \underline{63/67} & $69.50{\pm}12.36$ & $68.56{\pm}12.88$ & $70.03{\pm}12.98$ \\
\midrule
\multicolumn{9}{l}{\emph{Language-guided supervision}} \\
GAP~\cite{xiang_generative_2023}$^{\dagger}$ & $98.52{\pm}1.03$ & $98.52{\pm}1.03$ & $98.52{\pm}1.01$ & $\underline{91.77{\pm}5.60}$ & 62/67 & $\underline{86.11{\pm}6.18}$ & $\underline{85.40{\pm}7.13}$ & $\underline{86.70{\pm}6.08}$ \\

\textbf{PhysioAI} & $\mathbf{99.03{\pm}1.34}$ & $\mathbf{99.06{\pm}1.30}$ & $\mathbf{99.08{\pm}1.27}$ & $\mathbf{94.64{\pm}7.36}$ & \textbf{64/67} & $\mathbf{87.44{\pm}7.69}$ & $\mathbf{87.30{\pm}7.77}$ & $\mathbf{87.44{\pm}6.72}$ \\
\bottomrule
\end{tabular}
\\[-1pt]
\parbox{0.98\textwidth}{\scriptsize $^{\dagger}$Protocol-controlled reimplementation following the released GAP method and prompt structure; text assets were regenerated and frozen for this study.}
\end{table*}

\paragraph{Comparison with existing methods.} Table~\ref{tab:main} summarises the comparative evaluation results. PhysioAI achieved the highest observed value for every reported endpoint, although the margins varied. On KiMoRe Overall, the leading methods were already close to saturation: PhysioAI reached $99.03{\pm}1.34\%$, only 0.27 percentage points (pp) above MS-G3D. The larger observed margins were 2.87~pp on Hard-67 and 1.33~pp on UI-PRMD Overall. On UI-PRMD, PhysioAI's macro-F1 and balanced accuracy were also 1.90 and 0.74~pp higher than the strongest corresponding results. We attribute these improvements to CKD-guided semantic supervision, which encourages skeleton representations to retain rehabilitation-relevant kinematic cues not conveyed by categorical labels alone. This guidance may be particularly useful for fine-grained exercise classes and variable executions, consistent with the larger gains on Hard-67 and UI-PRMD. Section~\ref{sec:ablations} examines this interpretation through the description, projection, and alignment ablations.

GAP is the most relevant comparator because it also uses text-derived supervision during skeleton-model training. GAP relies on generated global and body-part action descriptions, whereas PhysioAI uses structured CKD anchors that explicitly encode movement region, movement plane, joint action, support pattern, laterality, and movement phase. Under the same evaluation protocol, PhysioAI obtained the highest observed results on Hard-67 and UI-PRMD. These findings indicate that structured clinical semantics can support language-guided physiotherapy action recognition while preserving competitive performance relative to a strong language-based baseline.


\subsection{Hard-67 results}
\begin{figure}[h]
    \centering
    \includegraphics[width=0.78\columnwidth]{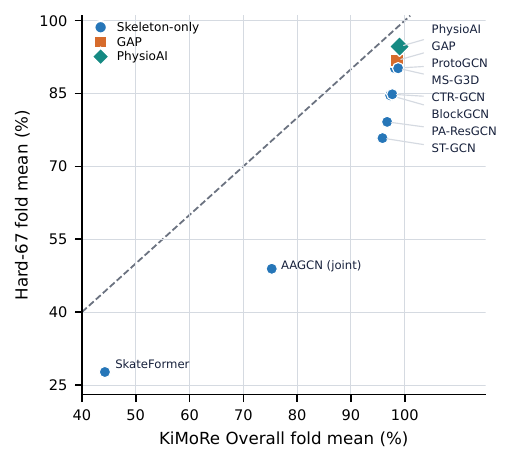}
    \caption{Cross-model Hard-67 stress-test profile. KiMoRe Overall fold-mean accuracy is plotted against Hard-67 fold-mean accuracy. The dashed identity line indicates equal performance on the two endpoints; points below the line show lower accuracy on Hard-67. All evaluated methods fall below the identity line, while PhysioAI shows the smallest observed reduction. }
    \label{fig:hard67_profile}
\end{figure}

The results in Table~\ref{tab:main} show that every evaluated method had a lower Hard-67 mean than its KiMoRe Overall, suggesting this subset remains challenging beyond the ST-GCN model used to construct it (Section~\ref{sec:hard67}). PhysioAI had the smallest observed reduction and reached $94.64{\pm}7.36\%$ on Hard-67, compared with $91.77{\pm}5.60\%$ for GAP. For the pooled endpoint, PhysioAI correctly classified 64/67 samples, compared with 63/67 for both MS-G3D and ProtoGCN, the joint second-best pooled results. Figure~\ref{fig:hard67_profile} visualises this cross-model pattern.

\begin{figure}[!t]
    \centering
    \includegraphics[width=0.78\columnwidth]{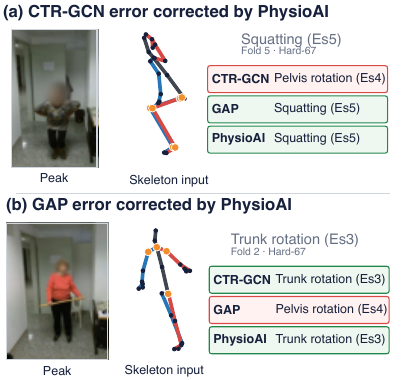}
    \caption{\textbf{Qualitative correction examples from Hard-67.} Each panel shows a face-obscured RGB frame, its frame-normalised skeleton, and model predictions. (a) CTR-GCN confuses Es5 squatting with pelvis rotation; (b) GAP confuses Es3 trunk rotation with pelvis rotation. PhysioAI is correct in both cases. Green and red boxes indicate correct and incorrect predictions. RGB frames are shown only for human interpretation.}
    \label{fig:hard67_examples}
\end{figure}

The Hard-67 result is consistent with the hypothesis that CKD guidance encourages the model to retain exercise-defining kinematic information rather than relying only on common execution patterns. The ablations in Section~\ref{sec:ablations} further show that performance varied with both the semantic descriptions and the alignment design. Figure~\ref{fig:hard67_examples} shows two selected Hard-67 cases misclassified by CTR-GCN or GAP but correctly classified by PhysioAI. Although these examples illustrate selected corrections, they do not establish causality or quantify their frequency.

\subsection{Ablation Studies}
\label{sec:ablations}

\paragraph{Contribution of semantic supervision and description design.} Table~\ref{tab:text} first compares PhysioAI with the same skeleton recogniser trained using CE alone, and then varies the semantic descriptions while holding the projection and alignment strategy fixed. Compared with classification-only training, the full CKD configuration improved KiMoRe Overall, Hard-67, and UI-PRMD Overall by 1.35, 9.82, and 1.59~pp, respectively. Class-name and generic descriptions also improved some endpoints, but neither matched the full CKD on UI-PRMD. Biomechanics-only descriptions achieved the highest KiMoRe and Hard-67 means, whereas the full CKD achieved the highest UI-PRMD mean. These results show that the complete CKD supervision improved the CE-only model across both datasets, while performance also depended on the content of the class descriptions.

\begin{table}[h]
\centering
\caption{Contribution of semantic supervision and semantic-description controls. The first row removes the semantic branch; the remaining rows use the same projection and alignment strategy with different text sources. Values are five-fold mean$\pm$sample SD (\%). Bold indicates the highest observed mean.}
\label{tab:text}
\scriptsize
\setlength{\tabcolsep}{2.5pt}
\begin{tabular}{lccc}
\toprule
Description & KiMoRe & Hard-67 & UI-PRMD \\
\midrule
No semantics (CE only) & $97.68{\pm}2.56$ & $84.82{\pm}24.02$ & $85.85{\pm}8.94$ \\
T0: class name & $98.76{\pm}1.81$ & $91.79{\pm}12.60$ & $85.64{\pm}7.87$ \\
T1: generic observable & $98.76{\pm}1.23$ & $91.79{\pm}12.60$ & $83.53{\pm}8.74$ \\
T2: biomechanics only & $\mathbf{99.29{\pm}1.01}$ & $\mathbf{97.50{\pm}5.59}$ & $85.41{\pm}7.16$ \\
T3: full CKD & $99.03{\pm}1.34$ & $94.64{\pm}7.36$ & $\mathbf{87.44{\pm}7.69}$ \\
\bottomrule
\end{tabular}
\end{table}

\paragraph{Effect of cross-modal projection.} This ablation tested whether the results depended on the fixed Gaussian projection or benefited from additional learnable cross-modal capacity. Table~\ref{tab:projection} compares fixed mappings, learned linear mappings, and a learned multilayer perceptron (MLP), while holding the CKD descriptions and training objective constant. The fixed Gaussian mapping achieved the most balanced results across the two datasets. The learned skeleton-to-CLIP linear mapping matched it on KiMoRe Overall and Hard-67 but was lower on UI-PRMD, while the other learned mappings were lower across all three endpoints. Additional projection capacity therefore provided no consistent advantage in this setting.

\begin{table}[h]
\centering
\caption{Cross-modal projection controls with the CKD descriptions and training objective held fixed. Values are five-fold mean$\pm$sample SD (\%). $T$ and $S$ denote the text and skeleton feature spaces. P0/P1 add no trainable parameters, P2/P3 add 131,072 parameters, and P4 adds 394,240 parameters. Bold indicates the highest observed mean, including ties.}
\label{tab:projection}
\scriptsize
\setlength{\tabcolsep}{2.0pt}
\begin{tabular}{lccc}
\toprule
Mapping & KiMoRe & Hard-67 & UI-PRMD \\
\midrule
P0: Gaussian $T{\rightarrow}S$
& $\mathbf{99.03{\pm}1.34}$
& $\mathbf{94.64{\pm}7.36}$
& $\mathbf{87.44{\pm}7.69}$ \\

P1: Orthogonal $T{\rightarrow}S$
& $98.23{\pm}1.75$
& $92.14{\pm}7.21$
& $85.23{\pm}7.88$ \\

P2: Linear $T{\rightarrow}S$
& $96.37{\pm}2.82$
& $86.13{\pm}11.03$
& $83.12{\pm}6.37$ \\

P3: Linear $S{\rightarrow}T$
& $\mathbf{99.03{\pm}1.34}$
& $\mathbf{94.64{\pm}7.36}$
& $84.51{\pm}7.29$ \\

P4: MLP $S{\rightarrow}T$
& $98.34{\pm}1.27$
& $90.32{\pm}6.46$
& $84.09{\pm}6.50$ \\
\bottomrule
\end{tabular}
\end{table}

\paragraph{Alignment weighting and onset.} This ablation tested whether adding semantic alignment was sufficient by itself, or whether its weighting and activation time affected the result. Table~\ref{tab:loss} compares fixed and uncertainty weighting with alignment activated at epoch~1 or epoch~6. Every alignment variant improved the observed Hard-67 mean over the classification-only model, but the Overall results depended on both weighting and onset. Fixed weighting did not improve UI-PRMD, while delayed uncertainty weighting was the only aligned configuration that exceeded the classification-only model on both Overall endpoints while matching the highest Hard-67 mean. The benefit therefore depended on how semantic alignment was incorporated, rather than on the presence of an auxiliary cosine loss alone.

\begin{table}[h]
\centering
\caption{Ablation of alignment weighting and onset. All aligned variants used cosine alignment with the same full CKD anchors and fixed Gaussian projection. Values are five-fold mean$\pm$sample SD (\%). Bold indicates the highest observed mean, including ties.}
\label{tab:loss}
\scriptsize
\setlength{\tabcolsep}{2.2pt}
\begin{tabular}{lccc}
\toprule
Configuration & KiMoRe & Hard-67 & UI-PRMD \\
\midrule
CE only
& $97.68{\pm}2.56$
& $84.82{\pm}24.02$
& $85.85{\pm}8.94$ \\

Fixed $\lambda{=}0.5$, epoch 1
& $98.49{\pm}1.68$
& $90.54{\pm}11.87$
& $84.55{\pm}8.47$ \\

Fixed $\lambda{=}0.5$, epoch 6
& $97.95{\pm}2.00$
& $87.68{\pm}17.84$
& $84.21{\pm}10.79$ \\

UW, epoch 1
& $\mathbf{99.29{\pm}1.01}$
& $\mathbf{94.64{\pm}7.36}$
& $83.73{\pm}10.04$ \\

UW, epoch 6 (PhysioAI)
& $99.03{\pm}1.34$
& $\mathbf{94.64{\pm}7.36}$
& $\mathbf{87.44{\pm}7.69}$ \\
\bottomrule
\end{tabular}
\end{table}

\section{Limitations and Future Work}
\label{sec:limitations}

This study used two relatively small public rehabilitation datasets and therefore cannot fully characterise variation across patient populations, functional abilities, and execution settings. Hard-67 revealed larger gains than the full KiMoRe benchmark but was derived from the training dynamics of a single reference model without clinician adjudication. It should therefore be viewed as a model-derived difficult-case benchmark rather than a universal definition of clinically challenging movement or evidence of broad robustness.

The implementation used CTR-GCN~\cite{chen_2021_channelwise} for its channel-wise topology refinement. However, PhysioAI is designed as training-time semantic supervision rather than a backbone-specific architecture: CKD alignment operates on pre-classification skeleton features and could be adapted to compatible encoders. Whether the benefits generalise remains to be established across a broader range of skeleton architectures.

PhysioAI recognises prescribed exercise identity but does not diagnose or assess movement quality, therapeutic correctness, or clinical outcomes. The current evaluation also does not establish performance across clinically defined movement difficulties. Future work will investigate clinician-guided hard-case datasets and multi-model difficulty criteria to better capture clinically meaningful movement challenges.
\section{Conclusion}
We proposed PhysioAI to address skeleton-based physiotherapy action recognition under limited training data, fine-grained class distinctions, and substantial execution variability. Results across KiMoRe, UI-PRMD, and Hard-67 show that CKD-guided semantic supervision can complement categorical labels, with its clearest observed advantages in the more challenging settings. The ablations show that these improvements do not arise from adding text or an auxiliary alignment loss alone; rather, they depend on how the semantic content is constructed and incorporated during training. PhysioAI therefore provides a practical approach for integrating structured rehabilitation knowledge into representation learning while retaining skeleton-only inference.
\FloatBarrier
{
    \small
    \bibliographystyle{ieeenat_fullname}
    \bibliography{main}
}

\end{document}


\maketitle
\thispagestyle{fancy}

This supplementary material provides Clinical Knowledge Dictionary (CKD) construction and sentence-selection details, experimental protocols, per-method settings, fold-level results, Hard-67 construction, and an index of the accompanying machine-readable assets.

\section{CKD Construction and Sentence Selection}
\label{sec:supp_ckd}

This section details the CKD construction procedure and the fixed semantic assets used in the reported experiments.

\subsection{Frozen exercise catalogue}

The catalogue contains 35 actions across five domains: functional movement screening (7 actions), lower-limb exercise (12), upper-limb exercise (7), balance and postural control (5), and gait or locomotion (4). For each action, \texttt{catalogue\_35.csv} records the fixed catalogue order, action identifier, display name, domain, laterality, public source hint, and brief kinematic definition. The source hints provide prompt context.

The complete pre-selection structured entries are provided in \texttt{structured\_entries\_35.json}. Each record links the structured entry to its catalogue position, selected candidate index, and final selected sentence:
\begin{center}
\small
\texttt{structured entry} $\rightarrow$ \texttt{six candidates} $\rightarrow$ \texttt{selected sentence}.
\end{center}

\subsection{Two-stage language generation}

Table~\ref{tab:supp_generation_settings} summarises the CKD language-generation settings.

\begin{table}[h]
\centering
\caption{Recorded CKD language-generation settings.}
\label{tab:supp_generation_settings}
\small
\begin{tabular}{lcc}
\toprule
Setting & Structured entry & Candidate sentences \\
\midrule
Recorded model alias & \texttt{gpt-4o} & \texttt{gpt-4o} \\
Temperature & 0.2 & 0.5 \\
Maximum output tokens & 3500 & 1200 \\
Candidates per action & -- & 6 \\
Maximum candidate length & -- & 40 words \\
\bottomrule
\end{tabular}
\end{table}

\paragraph{Stage 1: structured CKD entries.}
The first prompt applies a fixed six-step structured Chain-of-Thought (CoT) schema~\cite{wei_chain--thought_2022}: (0) movement taxonomy, (1) movement anatomy, (2) movement-category discriminators, (3) population-specific execution variations, (4) NTU-25 joint correspondence, and (5) phase-level movement landmarks. The system message, four few-shot exemplars, user template, catalogue fields, and additional constraints for the Contrastive Language--Image Pre-training (CLIP) model~\cite{radford_2021_learning} are provided in \texttt{stage1\_prompt.txt}. The 35 action-specific rendered prompts are provided in \texttt{stage1\_prompts\_35.json}.

\paragraph{Stage 2: discriminative candidates.}
For each action, the second prompt reads the movement taxonomy, reasoning summary, body-part descriptions, movement-category discriminators, and phase anchors from the frozen structured entry. It also supplies the action's predefined confusable pairs and action-family guidance. The prompt requests six declarative sentences of at most 40 words, prioritising the moving region, movement plane or axis, joint action, support pattern, laterality, and phases while excluding pathology wording and observer-facing instructions. The template is provided in \texttt{stage2\_prompt.txt}; all 35 rendered prompts are provided in \texttt{stage2\_prompts\_35.json}.

The complete 210-candidate bank is provided in \texttt{candidates\_210.csv}; each row records the action, candidate index, sentence, and whether it was retained. The final anchor sentences and selected indices are provided in \texttt{selected\_sentences\_35.csv}.

\subsection{CLIP encoding and fixed projection}

Candidate sentences were tokenised using the frozen \texttt{openai/clip-vit-base-patch32} text encoder with truncation to the encoder's 77-token limit. The resulting 512-dimensional text embeddings were $\ell_2$-normalised. A Gaussian matrix in $\mathbb{R}^{512\times256}$ was generated with seed 42 and normalised column-wise, following random-projection practice~\cite{bingham_random_2001}. Each text embedding was multiplied by this matrix and normalised again, yielding the 256-dimensional projected anchor used for sentence selection and later semantic supervision. The CLIP encoder, projection, and projected anchors remained fixed during skeleton-model training.

The complete KiMoRe~\cite{capecci_2019_the} and UI-PRMD~\cite{vakanski_data_2018} label-to-CKD mappings are provided in \texttt{class\_mappings.json}. KiMoRe maps five labels to five CKD actions, while UI-PRMD maps ten labels to ten CKD actions.

\subsection{Text-geometry objective}

Let $\mathcal{A}$ denote the projected 35-action anchor bank. Sentence selection minimises
\begin{equation}
\begin{split}
J(\mathcal{A})={}&
\mu_{\mathrm{all}}
+0.85\mu_{\mathrm{conf}}
+0.35\mu_{\mathrm{NN}}\\
&+0.50s_{\max}
+0.20\mu_{\mathrm{family}}
+0.30N_{>0.90},
\end{split}
\label{eq:supp_selection_objective}
\end{equation}
where $\mu_{\mathrm{all}}$ is the mean cosine similarity over all unordered anchor pairs; $\mu_{\mathrm{conf}}$ is the mean over 15 predefined confusable pairs; $\mu_{\mathrm{NN}}$ is the mean, across actions, of each action's maximum similarity to another anchor; $s_{\max}$ is the largest pairwise similarity; $\mu_{\mathrm{family}}$ is the mean similarity among within-family pairs; and $N_{>0.90}$ counts anchor pairs with cosine similarity above 0.90. Lower values indicate a more separated projected anchor bank.

The sentence-format penalty is added to the geometry objective when evaluating a candidate for the current catalogue coordinate. Table~\ref{tab:supp_format_penalty} gives the complete penalty definition.

\begin{table}[h]
\centering
\caption{Sentence-format penalty used during candidate selection.}
\label{tab:supp_format_penalty}
\small
\begin{tabular}{lc}
\toprule
Condition & Penalty \\
\midrule
Starts with a banned observer-facing expression & 1.00 \\
Does not contain the declarative label marker \texttt{action:} & 0.75 \\
Does not end with a period & 0.05 \\
\bottomrule
\end{tabular}
\end{table}

The banned starters were \texttt{imagine}, \texttt{notice}, \texttt{watch}, \texttt{observe}, \texttt{view}, \texttt{spot}, \texttt{examine}, \texttt{pay attention}, \texttt{witness}, and \texttt{assume}. The complete objective definition and final anchor-bank summary are provided in \texttt{selection\_\allowbreak objective.json}.

\subsection{Confusable pairs and action families}

The 15 predefined confusable pairs were:
\begin{enumerate}
\small
\item \texttt{trunk\_rotation} / \texttt{pelvis\_rotation};
\item \texttt{lateral\_trunk\_tilt} / \texttt{trunk\_rotation};
\item \texttt{lateral\_trunk\_tilt} / \texttt{pelvis\_rotation};
\item \texttt{shoulder\_abduction} / \texttt{shoulder\_scaption};
\item \texttt{shoulder\_abduction} / \texttt{shoulder\_flexion};
\item \texttt{shoulder\_flexion} / \texttt{reaching\_forward};
\item \texttt{shoulder\_extension} / \texttt{reaching\_forward};
\item \texttt{sit\_to\_stand} / \texttt{stand\_to\_sit};
\item \texttt{forward\_step\_up} / \texttt{lateral\_step\_up};
\item \texttt{stair\_ascent} / \texttt{stair\_descent};
\item \texttt{weight\_shifting\_lateral} / \texttt{weight\_shifting\_anteroposterior};
\item \texttt{deep\_squat} / \texttt{sit\_to\_stand};
\item \texttt{walking\_normal} / \texttt{timed\_up\_and\_go};
\item \texttt{hurdle\_step} / \texttt{forward\_step\_up}; and
\item \texttt{inline\_lunge} / \texttt{side\_lunge}.
\end{enumerate}
The same list is provided in \texttt{critical\_pairs\_15.json}.

The five action families were:
\begin{description}
\small
\item[Trunk/pelvis:] \texttt{lateral\_trunk\_tilt}, \texttt{trunk\_rotation}, \texttt{pelvis\_rotation}, and \texttt{trunk\_stability\_pushup}.
\item[Transfer:] \texttt{sit\_to\_stand}, \texttt{stand\_to\_sit}, \texttt{timed\_up\_and\_go}, and \texttt{sit\_reach\_forward}.
\item[Shoulder:] \texttt{shoulder\_mobility}, \texttt{shoulder\_abduction}, \texttt{shoulder\_flexion}, \texttt{shoulder\_extension}, \texttt{shoulder\_scaption}, \texttt{shoulder\_internal\_external\_rotation}, and \texttt{reaching\_forward}.
\item[Gait/step:] \texttt{walking\_normal}, \texttt{stair\_ascent}, \texttt{stair\_descent}, \texttt{forward\_step\_up}, \texttt{lateral\_step\_up}, and \texttt{stepping\_task}.
\item[Weight/balance:] \texttt{tandem\_stance}, \texttt{weight\_shifting\_lateral}, \texttt{weight\_shifting\_anteroposterior}, and \texttt{single\_leg\_stance}.
\end{description}
The machine-readable assignments are provided in \texttt{action\_families.json}.

\subsection{Greedy coordinate-descent selection}

Selection used the fixed 35-action catalogue order and did not use skeleton recognition accuracy:
\begin{enumerate}
\item For each action, form the set of candidates whose format penalty is below 0.5. If that set is empty, use all six candidates. Initialise the action with a minimum-penalty candidate.
\item Traverse actions in the fixed catalogue order.
\item At the current action, encode the complete current 35-sentence bank with frozen CLIP, apply the fixed projection, and calculate Eq.~\ref{eq:supp_selection_objective} plus the current sentence's format penalty.
\item Replace only the current action with each alternative candidate in turn. Candidates with format penalty at least 1.0 are not considered. Re-encode the complete proposal bank and calculate the combined score.
\item Retain an alternative only when its combined score is strictly lower than the best score found for that coordinate.
\item Repeat complete catalogue passes until no action sentence changes.
\end{enumerate}

The selected bank had a final geometry objective of 4.258028, mean pairwise cosine similarity of 0.797983, maximum pairwise similarity of 0.927219, mean nearest-neighbour similarity of 0.884768, mean confusable-pair similarity of 0.844969, mean within-family similarity of 0.842717, and six pairs above 0.90. These values describe the geometry of the final 35-action text-anchor bank; they are not recognition accuracies or candidate-level scores.

\subsection{Deep Squat construction example}

Table~\ref{tab:supp_deep_squat_candidates} gives all six candidates generated for \emph{Deep Squat}. Candidate 3 was retained by coordinate-descent selection based on the objective of the complete 35-action anchor bank.

\begin{table}[h]
\centering
\caption{Complete candidate set for \emph{Deep Squat}. The selected candidate is shown in bold.}
\label{tab:supp_deep_squat_candidates}
\small
\begin{tabularx}{\linewidth}{cX}
\toprule
Index & Candidate sentence \\
\midrule
1 & Deep squat action: Lower limbs flex in the sagittal plane as hips, knees, and ankles bend simultaneously, arms remain overhead. \\
2 & Deep squat action: Initiate with bilateral lower limb flexion in the sagittal plane, progressing to full descent with arms overhead. \\
\textbf{3} & \textbf{Deep squat action: Begin with sagittal plane flexion of hips, knees, and ankles, maintaining arms overhead for balance.} \\
4 & Deep squat action: Execute a full cycle of sagittal plane flexion and extension in lower limbs, with arms overhead. \\
5 & Deep squat action: Achieve maximum lower limb flexion in the sagittal plane, maintaining thoracic extension and arms overhead. \\
6 & Deep squat action: Perform bilateral lower limb flexion and extension in the sagittal plane, with stable arms overhead. \\
\bottomrule
\end{tabularx}
\end{table}
\section{Experimental Protocol and Reproducibility}
\label{sec:supp_protocol}

This section documents the fixed subject partitions, skeleton preparation, common loader settings, software environments, checkpoint convention, and metric calculations used for the results in the main paper. Machine-readable copies are supplied in \texttt{assets/protocol/}.

\subsection{Subject-disjoint five-fold evaluation}

All reported methods used the same five subject-disjoint folds. For a given fold, every sequence from its listed subjects formed the held-out set and all remaining subjects formed the training set. Thus, no subject contributed sequences to both partitions within a fold. The complete held-out subject lists are provided in \texttt{kimore\_5fold.json} and \texttt{uiprmd\_5fold.json}. Table~\ref{tab:supp_fold_counts} gives the corresponding fold sizes.

\begin{table}[H]
\centering
\caption{Fixed subject-disjoint partitions. Train and test are sequence counts.}
\label{tab:supp_fold_counts}
\small
\begin{tabular}{llrrr}
\toprule
Dataset & Fold & Test subjects & Train & Test \\
\midrule
KiMoRe & 1 & 18 & 309 & 91 \\
       & 2 & 17 & 313 & 87 \\
       & 3 & 15 & 325 & 75 \\
       & 4 & 14 & 328 & 72 \\
       & 5 & 14 & 325 & 75 \\
\midrule
UI-PRMD & 1 & 2 (S01--S02) & 1042 & 284 \\
        & 2 & 2 (S03--S04) & 1002 & 324 \\
        & 3 & 2 (S05--S06) & 1002 & 324 \\
        & 4 & 2 (S07--S08) & 1120 & 206 \\
        & 5 & 2 (S09--S10) & 1138 & 188 \\
\bottomrule
\end{tabular}
\end{table}

The prepared data contained 400 KiMoRe sequences from 78 subjects and 1,326 UI-PRMD sequences from ten subjects. Table~\ref{tab:supp_label_maps} records the label mappings used in evaluation. Tensor shapes and fold counts are supplied in \texttt{dataset\_protocol.json}.

\begin{table}[h]
\centering
\caption{Dataset class mappings used by the reported experiments.}
\label{tab:supp_label_maps}
\small
\begin{tabular}{lll}
\toprule
Dataset & Label/source code & Mapped CKD action \\
\midrule
KiMoRe & 0 / Es1 & Shoulder flexion \\
       & 1 / Es2 & Lateral trunk tilt \\
       & 2 / Es3 & Trunk rotation \\
       & 3 / Es4 & Pelvis rotation \\
       & 4 / Es5 & Deep squat \\
\midrule
UI-PRMD & 0 / m01 & Deep squat \\
        & 1 / m02 & Hurdle step \\
        & 2 / m03 & Inline lunge \\
        & 3 / m04 & Side lunge \\
        & 4 / m05 & Sit-to-stand \\
        & 5 / m06 & Active straight-leg raise \\
        & 6 / m07 & Shoulder abduction \\
        & 7 / m08 & Shoulder extension \\
        & 8 / m09 & Shoulder internal / external rotation \\
        & 9 / m10 & Shoulder scaption \\
\bottomrule
\end{tabular}
\end{table}

\subsection{Skeleton preparation and common loader}

Each prepared archive stores Cartesian skeleton coordinates as a tensor of shape $N\times300\times150$, where $150=2$ person slots $\times25$ joints $\times3$ coordinates. The flattened order is person, joint, then $(x,y,z)$ coordinate. KiMoRe contains one tracked person, so its second person slot is zero-padded. The loader reshapes each sequence to channel--time--joint--person order, $(3,T,25,2)$. The exact 25-joint order is provided in \texttt{ntu25\_joint\_order.json}, and the complete loader record is provided in \texttt{preprocessing\_and\_loader.json}. Dataset preparation linearly resampled each sequence to 300 frames; the loader introduced no additional spatial normalisation.

Temporal preparation then followed the relevant backbone. Spatial Temporal Graph Convolutional Network (ST-GCN)~\cite{yan_2018_spatial} inputs used automatic zero-padding when required and random temporal selection during training; evaluation used the leading model-length frames with zero-padding when required. The other common backbones used valid-frame crop-and-resize: training sampled a crop proportion uniformly from $[0.5,1.0]$ with a minimum of 64 frames, while evaluation used the centred 95\% interval. The selected crop was linearly interpolated to the backbone-specific temporal length. Those lengths and model-specific transformations, including Skeletal-Temporal Transformer (SkateFormer)~\cite{do_skateformer_2024} joint reordering and Prototypical Graph Convolutional Network (ProtoGCN)~\cite{liu_revealing_2025} clip protocol, are reported in Section~\ref{sec:supp_methods}.

Training-only rotations sampled each axis from $[-0.3,0.3]$ radians. Unless noted in Section~\ref{sec:supp_methods}, loaders used four workers, batch size 32, and \texttt{drop\_last} set to true for training and false for evaluation.

\subsection{Training determinism and environments}

The fixed training seed was 1 for Python, NumPy, PyTorch, and all CUDA devices. cuDNN deterministic execution was enabled and benchmarking was disabled. The independently fixed semantic-projection seed was 42; it is not an additional training repetition. Unless otherwise stated in the per-method records, training used 80 epochs, cross-entropy (CE), and Nesterov stochastic gradient descent (SGD) with momentum 0.9, weight decay $5\times10^{-4}$, and base learning rate 0.1. The learning rate was linearly warmed over five epochs and multiplied by 0.1 at epochs 35, 55, and 75.

Table~\ref{tab:supp_environments} gives the two software environments used. All runs used one NVIDIA GeForce RTX~3090 GPU with 24~GB memory. The settings are also provided in \texttt{reproducibility\_settings.json}.

\begin{table}[h]
\centering
\caption{Software and hardware environments.}
\label{tab:supp_environments}
\small
\begin{tabular}{lcc}
\toprule
Component & Most experiments & ProtoGCN \\
\midrule
Python & 3.10.19 & 3.7.11 \\
PyTorch & 2.5.1 & 1.11.0 \\
CUDA & 12.1 & 11.3 \\
cuDNN & 9.1 & 8.2 \\
GPU & RTX 3090, 24~GB & RTX 3090, 24~GB \\
\bottomrule
\end{tabular}
\end{table}

\subsection{Checkpoint-selection convention}

For every reported method and fold, the held-out subject fold was evaluated after every training epoch. Among epochs 1--80, we retained the earliest checkpoint attaining the highest held-out Overall accuracy. Operationally, a checkpoint was replaced only after a strict accuracy improvement, which implements the earliest-epoch tie break. This is a held-out-fold test-best convention and should not be interpreted as evaluating the test fold only once.

Hard-67 membership, labels, and accuracy did not enter checkpoint selection. Each Hard-67 sequence was evaluated using the already selected Overall-accuracy checkpoint for its subject's held-out fold. The same rule is recorded in \texttt{checkpoint\_selection.txt}.

\subsection{Metric definitions and aggregation}

Let $n$ denote the number of held-out sequences in a fold. Overall accuracy is
\begin{equation}
\mathrm{Overall}=\frac{1}{n}\sum_{i=1}^{n}\mathbf{1}(\hat{y}_i=y_i).
\end{equation}
For each true class $c$ represented in that fold, precision, recall, and F1 are
\begin{equation}
\begin{aligned}
P_c&=\frac{TP_c}{TP_c+FP_c}, &
R_c&=\frac{TP_c}{TP_c+FN_c},\\
F1_c&=\frac{2P_cR_c}{P_c+R_c}.&&
\end{aligned}
\end{equation}
with a zero value when the corresponding precision or F1 denominator is zero. Macro-F1 and balanced accuracy are the unweighted means
\begin{equation}
\mathrm{mF1}=\frac{1}{|\mathcal{C}_f|}\sum_{c\in\mathcal{C}_f}F1_c,
\qquad
\mathrm{BA}=\frac{1}{|\mathcal{C}_f|}\sum_{c\in\mathcal{C}_f}R_c,
\end{equation}
where $\mathcal{C}_f$ is the set of true classes present in held-out fold $f$. UI-PRMD fold 5 has no true label-9 sequence, so its fold-level mF1 and BA average the nine represented classes rather than inserting an artificial tenth-class value.

Each metric was first computed separately for the five folds. If $m_f$ is a fold value, the reported summary is the unweighted mean $\bar{m}=\frac{1}{5}\sum_{f=1}^{5}m_f$ and sample standard deviation
\begin{equation}
s=\sqrt{\frac{1}{4}\sum_{f=1}^{5}(m_f-\bar{m})^2}.
\end{equation}
These SD values describe variation among five subject folds from one fixed training seed; they do not estimate multi-seed training uncertainty. For pooled out-of-fold analyses, the five mutually exclusive held-out prediction sets were concatenated, so every sequence contributed exactly once. Hard-67 additionally reports the pooled number correct out of 67. The same definitions are provided in \texttt{metric\_definitions.txt}.
\section{Per-Method Implementation Details}
\label{sec:supp_methods}

This section records the final configurations used for the methods and ablations reported in the main paper. It distinguishes the shared comparison protocol from model-specific exceptions. Machine-readable copies are provided in \texttt{assets/methods/}.

The comparison set additionally includes attention-enhanced adaptive graph convolutional network (AAGCN)~\cite{shi_multistream_2020}, Multi-Scale 3D Graph Convolutional Network (MS-G3D)~\cite{liu_disentangling_2020}, Part-wise Attention Residual GCN (PA-ResGCN)~\cite{song_stronger_2020}, Channel-wise Topology Refinement Graph Convolution (CTR-GCN)~\cite{chen_2021_channelwise}, BlockGCN~\cite{zhou_blockgcn_2024}, and Generative Action-description Prompts (GAP)~\cite{xiang_generative_2023}.

\subsection{Common comparison protocol}

Every reported method used joint-coordinate input and was trained from scratch without pretrained action-recognition checkpoints. Sampling-derived temporal indices required by SkateFormer are not an additional sensing modality. Unless stated below, all methods followed the shared protocol in Section~\ref{sec:supp_protocol}.

Table~\ref{tab:supp_common_recipe} summarises the shared settings and documented exceptions. Model-specific temporal lengths and transformations are given below.

\begin{table}[h]
\centering
\caption{Controlled comparison recipe and model-specific exceptions.}
\label{tab:supp_common_recipe}
\small
\begin{tabularx}{\linewidth}{lX}
\toprule
Setting & Final configuration \\
\midrule
Input and initialisation & Joint coordinates; training from scratch; no pretrained action-recognition checkpoint. \\
Training length & 80 epochs, batch size 32, fixed training seed 1. \\
Common optimiser & Nesterov SGD, momentum 0.9, base learning rate 0.1, weight decay $5\times10^{-4}$. \\
Common schedule & Five-epoch linear warm-up; learning rate multiplied by 0.1 at epochs 35, 55, and 75. \\
Common augmentation & Training-only random three-axis rotation; model-length temporal crop and interpolation. \\
Evaluation & Five subject-disjoint folds; every epoch; earliest maximum held-out Overall checkpoint. \\
PA-ResGCN exception & Weight decay $2\times10^{-4}$. \\
ProtoGCN exceptions & Base learning rate 0.05, cosine annealing to zero, rotation bound $0.2$ radians, 100-frame clip protocol, and a separate software environment. \\
GAP and PhysioAI & Retained the common optimiser schedule but added their respective language or semantic-supervision objectives. \\
\bottomrule
\end{tabularx}
\end{table}

\subsection{Final method configurations}

\paragraph{ST-GCN, MS-G3D, PA-ResGCN, and CTR-GCN.}
ST-GCN used the joint stream with 300-frame inputs. MS-G3D used 50-frame inputs, 13 graph-convolution scales, and six G3D scales. PA-ResGCN used the \texttt{pa-resgcn-b19-r4} joint-stream model with 64-frame inputs and the weight-decay exception in Table~\ref{tab:supp_common_recipe}. The plain CTR-GCN baseline used 64-frame joint inputs and produced a 256-dimensional pre-classifier feature. All four otherwise followed the common recipe.

\paragraph{AAGCN.}
AAGCN used 64-frame joint inputs with 25 joints, two person slots, adaptive topology enabled, attention enabled, and zero dropout. Its learnable adjacency tensors were frozen in epochs 1--6 and trainable from epoch 7. The remaining optimiser and evaluation settings followed the common controlled protocol rather than a dataset-specific native recipe.

\paragraph{BlockGCN.}
BlockGCN used 64-frame, 25-joint inputs with the spatial graph, adaptive graph processing enabled, \texttt{alpha=false}, and zero dropout. The topology branch received raw joint coordinates. For the graph-convolution branch, coordinates were centred on one-based joint 21 while that joint's original trajectory was restored, matching the reported adapter. The final environment used \texttt{torch-topological}~0.1.7 and \texttt{einops}~0.6.1. Optimisation followed the common recipe.

\paragraph{SkateFormer.}
SkateFormer used the small configuration with stage depths $2/2/2/2$, channels $96/192/192/192$, embedding dimension 96, 32 heads, kernel size 7, attention dropout 0.5, stochastic-depth rate 0.2, and multilayer perceptron (MLP) ratio 4. The four temporal--joint partitions were $(8,8)$, $(8,12)$, $(8,8)$, and $(8,12)$. Before the model, one-based joint 21 was removed and the remaining 24 joints were reordered according to the official NTU partition. The exact zero-based order is provided in \texttt{method\_\allowbreak recipes.json}. Inputs were 64 frames, and sampling-derived temporal indices supplied the temporal positional encoding. Training used the common controlled recipe rather than SkateFormer's native dataset recipe.

\paragraph{ProtoGCN.}
ProtoGCN used joint features, 400 prototypes, eight graph filters, a 384-dimensional classifier input, and classifier weight 0.2. Spine alignment in \texttt{PreNormalize3D} was disabled. Training and validation used uniformly sampled 100-frame clips; validation used one clip per sequence, whereas test-time prediction averaged ten 100-frame clips. The batch size was 32. Nesterov SGD used base learning rate 0.05, momentum 0.9, weight decay $5\times10^{-4}$, and iteration-wise cosine annealing to zero over 80 epochs. Its software environment is listed in Table~\ref{tab:supp_environments}.

\paragraph{GAP.}
GAP used its four-part CTR-GCN joint-stream backbone with 64-frame inputs. For each rehabilitation action, we regenerated and then froze ten synonyms plus atomic descriptions for the head, hand, arm, hip, leg, and foot; the frozen KiMoRe and UI-PRMD assets are supplied as \texttt{gap\_kimore.json} and \texttt{gap\_uiprmd.json}. These assets were generated with \texttt{gpt-4o-2024-08-06} at temperature 0 using the released GAP prompt structure and did not use the PhysioAI CKD.

Following the released four-part structure, five text groups were constructed: the global action; global plus head; global plus hand and arm; global plus hip; and global plus leg and foot. A trainable CLIP ViT-B/32 text encoder was kept in float32 and used the same learning-rate scale as the skeleton optimiser. The final objective was
\begin{equation}
\mathcal{L}_{\mathrm{GAP}}=\mathcal{L}_{\mathrm{CE}}+0.8\mathcal{L}_{\mathrm{language}},
\end{equation}
with graph-hop parameter $k=8$. Text processing and language losses were used only during training; reported inference used the skeleton network and its classifier.

\paragraph{PhysioAI.}
PhysioAI used the same 64-frame CTR-GCN joint-stream backbone and 256-dimensional pre-classifier feature as the plain CTR-GCN comparison, together with the fixed CKD anchors and projection described in Section~\ref{sec:supp_ckd}. Training used class-matched cosine alignment and delayed uncertainty weighting (UW)~\cite{kendall_multi-task_2018}, with two learned log-scale parameters initialised to zero. Epochs 1--5 excluded the alignment loss and its log-scale term; the complete classification and alignment objective was active from epoch 6. At inference, the semantic assets, projection, and weighting parameters were discarded, leaving only CTR-GCN and its linear classifier.

\subsection{Ablation configurations}

All ablations used the CTR-GCN joint stream, the shared batch-32 protocol, a 256-dimensional skeleton feature, and 512-dimensional frozen CLIP text embeddings. The exact configuration records are in \texttt{ablation\_\allowbreak configurations.json}.

\paragraph{Semantic-description controls.}
T0 encoded only the action name in the template ``A skeleton sequence of a person performing [action].'' T1 used a generic observable action description without the structured CKD fields. T2 retained biomechanics-focused movement content, and T3 used the selected full CKD sentence. Projection and loss scheduling were fixed to P0 and delayed uncertainty weighting for all four variants. The exact 35 frozen sentences for every variant are supplied in \texttt{ablation\_text\_descriptions.json}.

\begin{table}[h]
\centering
\caption{Projection controls. $T$ is the 512-dimensional text space and $S$ is the 256-dimensional skeleton-feature space.}
\label{tab:supp_projection_configs}
\small
\begin{tabular}{lllr}
\toprule
ID & Direction & Mapping & Trainable parameters \\
\midrule
P0 & $T\rightarrow S$ & Fixed column-normalised Gaussian matrix & 0 \\
P1 & $T\rightarrow S$ & Fixed orthogonal matrix from reduced QR & 0 \\
P2 & $T\rightarrow S$ & Learned bias-free linear layer & 131,072 \\
P3 & $S\rightarrow T$ & Learned bias-free linear layer & 131,072 \\
P4 & $S\rightarrow T$ & Learned $256\rightarrow512\rightarrow512$ MLP with GELU & 394,240 \\
\bottomrule
\end{tabular}
\end{table}

The projection study held T3 descriptions and delayed uncertainty weighting fixed. P0 and P1 used seed 42 and introduced no trainable projection parameters. P2--P4 were optimised jointly with the skeleton network.

\paragraph{Alignment weighting and onset.}
Let
\begin{align}
U_{\mathrm{CE}}&=e^{-2s_{\mathrm{ce}}}\mathcal{L}_{\mathrm{CE}}+s_{\mathrm{ce}},\\
U_{\mathrm{align}}&=e^{-2s_{\mathrm{align}}}\mathcal{L}_{\mathrm{align}}+s_{\mathrm{align}}.
\end{align}
where both log-scale parameters are initialised to zero. Table~\ref{tab:supp_loss_schedules} states the exact active objective in both epoch ranges. All aligned variants used T3 descriptions, P0, and the same cosine-alignment loss.

\begin{table}[h]
\centering
\caption{Exact objective schedules in the weighting/onset ablation.}
\label{tab:supp_loss_schedules}
\small
\begin{tabularx}{\linewidth}{lXX}
\toprule
Configuration & Epochs 1--5 & Epochs 6--80 \\
\midrule
CE only & $\mathcal{L}_{\mathrm{CE}}$ & $\mathcal{L}_{\mathrm{CE}}$ \\
Fixed $\lambda=0.5$, epoch 1 & $\mathcal{L}_{\mathrm{CE}}+0.5\mathcal{L}_{\mathrm{align}}$ & Same \\
Fixed $\lambda=0.5$, epoch 6 & $\mathcal{L}_{\mathrm{CE}}$ & $\mathcal{L}_{\mathrm{CE}}+0.5\mathcal{L}_{\mathrm{align}}$ \\
UW, epoch 1 & $U_{\mathrm{CE}}+U_{\mathrm{align}}$ & Same \\
UW, epoch 6 (PhysioAI) & $U_{\mathrm{CE}}$; alignment branch excluded & $U_{\mathrm{CE}}+U_{\mathrm{align}}$ \\
\bottomrule
\end{tabularx}
\end{table}

For delayed uncertainty weighting, epochs 1--5 therefore optimise the uncertainty-weighted classification branch, not an alignment term with a zero-valued loss: both $\mathcal{L}_{\mathrm{align}}$ and $s_{\mathrm{align}}$ are absent until epoch 6. The corresponding configurations are provided in \texttt{assets/methods/}.
\section{Detailed Ablation and Fold-Level Results}
\label{sec:supp_fold_results}

This section provides the five fold values underlying the main comparison and ablation tables. Folds F1--F5 and aggregation follow Section~\ref{sec:supp_protocol}. Unless a cell is written as a count, values are percentages. Hard-67 fold cells report correct/total, while the aggregate reports the five-fold mean and sample SD followed by the pooled count. The complete values at six-decimal precision are supplied in \texttt{assets/results/}.

The machine-readable files preserve full fold precision and include the integer correct and test counts used for Overall accuracy. The bold and underline used in the main paper mark descriptive best and second-best observed endpoint values only and do not indicate statistical significance.

\begin{table}[!h]
\centering
\caption{Fold-level KiMoRe Overall accuracy and Hard-67 results for the main comparison. For Hard-67, the aggregate column gives fold mean$\pm$sample SD followed by the pooled correct/67 count.}
\label{tab:supp_main_kimore_hard}
\scriptsize
\setlength{\tabcolsep}{3.2pt}
\begin{tabular}{llrrrrrc}
\toprule
Method & Endpoint & F1 & F2 & F3 & F4 & F5 & Aggregate \\
\midrule
ST-GCN & KiMoRe Overall & 96.70 & 98.85 & 94.67 & 93.06 & 96.00 & $95.86{\pm}2.18$ \\
 & Hard-67 & 13/16 & 12/13 & 12/16 & 11/15 & 4/7 & $75.81{\pm}12.82$; 52/67 \\
\addlinespace[1pt]
AAGCN (joint) & KiMoRe Overall & 90.11 & 75.86 & 78.67 & 59.72 & 72.00 & $75.27{\pm}11.01$ \\
 & Hard-67 & 10/16 & 6/13 & 12/16 & 7/15 & 1/7 & $48.92{\pm}22.79$; 36/67 \\
\addlinespace[1pt]
MS-G3D & KiMoRe Overall & 97.80 & 100.00 & 100.00 & 100.00 & 96.00 & $98.76{\pm}1.81$ \\
 & Hard-67 & 15/16 & 13/13 & 16/16 & 15/15 & 4/7 & $90.18{\pm}18.66$; 63/67 \\
\addlinespace[1pt]
PA-ResGCN & KiMoRe Overall & 97.80 & 96.55 & 97.33 & 97.22 & 94.67 & $96.72{\pm}1.23$ \\
 & Hard-67 & 14/16 & 10/13 & 14/16 & 13/15 & 4/7 & $79.15{\pm}13.09$; 55/67 \\
\addlinespace[1pt]
CTR-GCN & KiMoRe Overall & 97.80 & 100.00 & 98.67 & 98.61 & 93.33 & $97.68{\pm}2.56$ \\
 & Hard-67 & 14/16 & 13/13 & 15/16 & 15/15 & 3/7 & $84.82{\pm}24.02$; 60/67 \\
\addlinespace[1pt]
BlockGCN & KiMoRe Overall & 96.70 & 97.70 & 97.33 & 100.00 & 94.67 & $97.28{\pm}1.92$ \\
 & Hard-67 & 14/16 & 11/13 & 15/16 & 15/15 & 4/7 & $84.60{\pm}16.46$; 59/67 \\
\addlinespace[1pt]
SkateFormer & KiMoRe Overall & 29.67 & 34.48 & 54.67 & 30.56 & 72.00 & $44.28{\pm}18.53$ \\
 & Hard-67 & 3/16 & 2/13 & 8/16 & 6/15 & 1/7 & $27.68{\pm}16.28$; 20/67 \\
\addlinespace[1pt]
ProtoGCN & KiMoRe Overall & 98.90 & 100.00 & 98.67 & 97.22 & 96.00 & $98.16{\pm}1.56$ \\
 & Hard-67 & 15/16 & 13/13 & 16/16 & 15/15 & 4/7 & $90.18{\pm}18.66$; 63/67 \\
\addlinespace[1pt]
GAP & KiMoRe Overall & 97.80 & 98.85 & 100.00 & 98.61 & 97.33 & $98.52{\pm}1.03$ \\
 & Hard-67 & 14/16 & 12/13 & 16/16 & 14/15 & 6/7 & $91.77{\pm}5.60$; 62/67 \\
\addlinespace[1pt]
PhysioAI & KiMoRe Overall & 97.80 & 100.00 & 100.00 & 100.00 & 97.33 & $99.03{\pm}1.34$ \\
 & Hard-67 & 14/16 & 13/13 & 16/16 & 15/15 & 6/7 & $94.64{\pm}7.36$; 64/67 \\
\bottomrule
\end{tabular}
\end{table}

\begin{table}[!h]
\centering
\caption{Fold-level UI-PRMD results for the main comparison. Overall, macro-F1 (mF1), and balanced accuracy (BA) are calculated independently in each held-out fold.}
\label{tab:supp_main_uiprmd}
\scriptsize
\setlength{\tabcolsep}{3.2pt}
\begin{tabular}{llrrrrrc}
\toprule
Method & Endpoint & F1 & F2 & F3 & F4 & F5 & Mean$\pm$SD \\
\midrule
ST-GCN & Overall & 84.51 & 85.49 & 85.19 & 80.10 & 87.23 & $84.50{\pm}2.66$ \\
 & mF1 & 78.88 & 82.78 & 85.28 & 75.08 & 88.10 & $82.02{\pm}5.15$ \\
 & BA & 81.11 & 83.07 & 86.70 & 77.75 & 87.14 & $83.15{\pm}3.93$ \\
\addlinespace[1pt]
AAGCN (joint) & Overall & 70.07 & 79.01 & 75.62 & 82.52 & 90.96 & $79.64{\pm}7.82$ \\
 & mF1 & 65.69 & 78.81 & 75.34 & 80.81 & 91.08 & $78.35{\pm}9.19$ \\
 & BA & 69.68 & 78.85 & 77.76 & 80.78 & 91.99 & $79.81{\pm}8.02$ \\
\addlinespace[1pt]
MS-G3D & Overall & 79.93 & 82.41 & 82.10 & 85.92 & 94.68 & $85.01{\pm}5.82$ \\
 & mF1 & 76.36 & 82.28 & 82.06 & 85.82 & 94.08 & $84.12{\pm}6.52$ \\
 & BA & 78.50 & 82.29 & 83.41 & 86.24 & 93.96 & $84.88{\pm}5.78$ \\
\addlinespace[1pt]
PA-ResGCN & Overall & 84.51 & 71.60 & 80.86 & 78.64 & 82.98 & $79.72{\pm}5.05$ \\
 & mF1 & 77.08 & 68.96 & 81.41 & 78.14 & 84.15 & $77.95{\pm}5.75$ \\
 & BA & 79.34 & 71.06 & 81.35 & 78.17 & 82.99 & $78.58{\pm}4.60$ \\
\addlinespace[1pt]
CTR-GCN & Overall & 79.58 & 83.33 & 75.93 & 94.66 & 95.74 & $85.85{\pm}8.94$ \\
 & mF1 & 77.72 & 83.39 & 75.29 & 94.73 & 95.85 & $85.40{\pm}9.51$ \\
 & BA & 78.68 & 83.33 & 78.65 & 94.41 & 96.23 & $86.26{\pm}8.51$ \\
\addlinespace[1pt]
BlockGCN & Overall & 63.73 & 45.68 & 70.37 & 65.53 & 86.17 & $66.30{\pm}14.52$ \\
 & mF1 & 61.40 & 41.14 & 67.52 & 61.72 & 82.46 & $62.85{\pm}14.84$ \\
 & BA & 64.98 & 43.40 & 67.61 & 64.74 & 84.20 & $64.98{\pm}14.51$ \\
\addlinespace[1pt]
SkateFormer & Overall & 76.06 & 22.22 & 43.21 & 62.62 & 55.85 & $51.99{\pm}20.43$ \\
 & mF1 & 71.43 & 10.07 & 41.02 & 64.38 & 48.42 & $47.07{\pm}23.98$ \\
 & BA & 75.00 & 21.67 & 43.37 & 66.11 & 53.03 & $51.83{\pm}20.76$ \\
\addlinespace[1pt]
ProtoGCN & Overall & 74.30 & 78.40 & 49.07 & 66.99 & 78.72 & $69.50{\pm}12.36$ \\
 & mF1 & 68.50 & 77.05 & 47.65 & 68.64 & 80.97 & $68.56{\pm}12.88$ \\
 & BA & 72.78 & 77.17 & 48.43 & 69.54 & 82.21 & $70.03{\pm}12.98$ \\
\addlinespace[1pt]
GAP & Overall & 86.27 & 87.04 & 79.94 & 81.55 & 95.74 & $86.11{\pm}6.18$ \\
 & mF1 & 86.68 & 86.90 & 78.62 & 78.87 & 95.94 & $85.40{\pm}7.13$ \\
 & BA & 88.06 & 87.12 & 80.88 & 81.50 & 95.93 & $86.70{\pm}6.08$ \\
\addlinespace[1pt]
PhysioAI & Overall & 87.32 & 83.64 & 77.16 & 91.75 & 97.34 & $87.44{\pm}7.69$ \\
 & mF1 & 84.86 & 83.80 & 77.78 & 92.45 & 97.59 & $87.30{\pm}7.77$ \\
 & BA & 84.79 & 84.13 & 79.86 & 91.60 & 96.83 & $87.44{\pm}6.72$ \\
\bottomrule
\end{tabular}
\end{table}

\begin{table}[!h]
\centering
\caption{Fold-level semantic-description ablation. Hard-67 fold entries are correct/total; their aggregate includes fold accuracy mean$\pm$sample SD and pooled correct/67.}
\label{tab:supp_text_ablation}
\scriptsize
\setlength{\tabcolsep}{3.0pt}
\begin{tabular}{llrrrrrc}
\toprule
Description & Endpoint & F1 & F2 & F3 & F4 & F5 & Aggregate \\
\midrule
No semantics (CE only) & KiMoRe Overall & 97.80 & 100.00 & 98.67 & 98.61 & 93.33 & $97.68{\pm}2.56$ \\
 & Hard-67 & 14/16 & 13/13 & 15/16 & 15/15 & 3/7 & $84.82{\pm}24.02$; 60/67 \\
 & UI-PRMD Overall & 79.58 & 83.33 & 75.93 & 94.66 & 95.74 & $85.85{\pm}8.94$ \\
\addlinespace[1pt]
T0: class name & KiMoRe Overall & 97.80 & 100.00 & 100.00 & 100.00 & 96.00 & $98.76{\pm}1.81$ \\
 & Hard-67 & 14/16 & 13/13 & 16/16 & 15/15 & 5/7 & $91.79{\pm}12.60$; 63/67 \\
 & UI-PRMD Overall & 77.46 & 83.02 & 81.48 & 88.35 & 97.87 & $85.64{\pm}7.87$ \\
\addlinespace[1pt]
T1: generic observable & KiMoRe Overall & 97.80 & 100.00 & 98.67 & 100.00 & 97.33 & $98.76{\pm}1.23$ \\
 & Hard-67 & 14/16 & 13/13 & 16/16 & 15/15 & 5/7 & $91.79{\pm}12.60$; 63/67 \\
 & UI-PRMD Overall & 77.46 & 81.79 & 74.69 & 86.89 & 96.81 & $83.53{\pm}8.74$ \\
\addlinespace[1pt]
T2: biomechanics only & KiMoRe Overall & 97.80 & 100.00 & 100.00 & 100.00 & 98.67 & $99.29{\pm}1.01$ \\
 & Hard-67 & 14/16 & 13/13 & 16/16 & 15/15 & 7/7 & $97.50{\pm}5.59$; 65/67 \\
 & UI-PRMD Overall & 81.69 & 84.26 & 76.54 & 89.32 & 95.21 & $85.41{\pm}7.16$ \\
\addlinespace[1pt]
T3: full CKD & KiMoRe Overall & 97.80 & 100.00 & 100.00 & 100.00 & 97.33 & $99.03{\pm}1.34$ \\
 & Hard-67 & 14/16 & 13/13 & 16/16 & 15/15 & 6/7 & $94.64{\pm}7.36$; 64/67 \\
 & UI-PRMD Overall & 87.32 & 83.64 & 77.16 & 91.75 & 97.34 & $87.44{\pm}7.69$ \\
\bottomrule
\end{tabular}
\end{table}

\begin{table}[!h]
\centering
\caption{Fold-level cross-modal projection ablation. $T$ and $S$ denote the text and skeleton feature spaces.}
\label{tab:supp_projection_ablation}
\scriptsize
\setlength{\tabcolsep}{3.0pt}
\begin{tabular}{llrrrrrc}
\toprule
Mapping & Endpoint & F1 & F2 & F3 & F4 & F5 & Aggregate \\
\midrule
P0: Gaussian $T{\rightarrow}S$ & KiMoRe Overall & 97.80 & 100.00 & 100.00 & 100.00 & 97.33 & $99.03{\pm}1.34$ \\
 & Hard-67 & 14/16 & 13/13 & 16/16 & 15/15 & 6/7 & $94.64{\pm}7.36$; 64/67 \\
 & UI-PRMD Overall & 87.32 & 83.64 & 77.16 & 91.75 & 97.34 & $87.44{\pm}7.69$ \\
\addlinespace[1pt]
P1: Orthogonal $T{\rightarrow}S$ & KiMoRe Overall & 97.80 & 100.00 & 96.00 & 100.00 & 97.33 & $98.23{\pm}1.75$ \\
 & Hard-67 & 14/16 & 13/13 & 14/16 & 15/15 & 6/7 & $92.14{\pm}7.21$; 62/67 \\
 & UI-PRMD Overall & 76.76 & 86.73 & 78.09 & 88.83 & 95.74 & $85.23{\pm}7.88$ \\
\addlinespace[1pt]
P2: Linear $T{\rightarrow}S$ & KiMoRe Overall & 92.31 & 97.70 & 100.00 & 95.83 & 96.00 & $96.37{\pm}2.82$ \\
 & Hard-67 & 13/16 & 11/13 & 16/16 & 14/15 & 5/7 & $86.13{\pm}11.03$; 59/67 \\
 & UI-PRMD Overall & 85.56 & 77.47 & 80.86 & 78.64 & 93.09 & $83.12{\pm}6.37$ \\
\addlinespace[1pt]
P3: Linear $S{\rightarrow}T$ & KiMoRe Overall & 97.80 & 100.00 & 100.00 & 100.00 & 97.33 & $99.03{\pm}1.34$ \\
 & Hard-67 & 14/16 & 13/13 & 16/16 & 15/15 & 6/7 & $94.64{\pm}7.36$; 64/67 \\
 & UI-PRMD Overall & 77.82 & 81.79 & 80.25 & 86.41 & 96.28 & $84.51{\pm}7.29$ \\
\addlinespace[1pt]
P4: MLP $S{\rightarrow}T$ & KiMoRe Overall & 97.80 & 96.55 & 98.67 & 100.00 & 98.67 & $98.34{\pm}1.27$ \\
 & Hard-67 & 14/16 & 11/13 & 15/16 & 15/15 & 6/7 & $90.32{\pm}6.46$; 61/67 \\
 & UI-PRMD Overall & 77.11 & 83.02 & 80.25 & 85.92 & 94.15 & $84.09{\pm}6.50$ \\
\bottomrule
\end{tabular}
\end{table}

\begin{table}[!h]
\centering
\caption{Fold-level alignment-weighting and onset ablation. UW denotes uncertainty weighting.}
\label{tab:supp_loss_ablation}
\scriptsize
\setlength{\tabcolsep}{3.0pt}
\begin{tabular}{llrrrrrc}
\toprule
Configuration & Endpoint & F1 & F2 & F3 & F4 & F5 & Aggregate \\
\midrule
CE only & KiMoRe Overall & 97.80 & 100.00 & 98.67 & 98.61 & 93.33 & $97.68{\pm}2.56$ \\
 & Hard-67 & 14/16 & 13/13 & 15/16 & 15/15 & 3/7 & $84.82{\pm}24.02$; 60/67 \\
 & UI-PRMD Overall & 79.58 & 83.33 & 75.93 & 94.66 & 95.74 & $85.85{\pm}8.94$ \\
\addlinespace[1pt]
Fixed $\lambda{=}0.5$, epoch 1 & KiMoRe Overall & 97.80 & 100.00 & 98.67 & 100.00 & 96.00 & $98.49{\pm}1.68$ \\
 & Hard-67 & 14/16 & 13/13 & 15/16 & 15/15 & 5/7 & $90.54{\pm}11.87$; 62/67 \\
 & UI-PRMD Overall & 74.30 & 82.10 & 81.17 & 88.35 & 96.81 & $84.55{\pm}8.47$ \\
\addlinespace[1pt]
Fixed $\lambda{=}0.5$, epoch 6 & KiMoRe Overall & 97.80 & 100.00 & 98.67 & 98.61 & 94.67 & $97.95{\pm}2.00$ \\
 & Hard-67 & 14/16 & 13/13 & 15/16 & 15/15 & 4/7 & $87.68{\pm}17.84$; 61/67 \\
 & UI-PRMD Overall & 72.89 & 80.25 & 76.85 & 93.20 & 97.87 & $84.21{\pm}10.79$ \\
\addlinespace[1pt]
UW, epoch 1 & KiMoRe Overall & 97.80 & 100.00 & 100.00 & 100.00 & 98.67 & $99.29{\pm}1.01$ \\
 & Hard-67 & 14/16 & 13/13 & 16/16 & 15/15 & 6/7 & $94.64{\pm}7.36$; 64/67 \\
 & UI-PRMD Overall & 71.13 & 86.11 & 77.16 & 86.89 & 97.34 & $83.73{\pm}10.04$ \\
\addlinespace[1pt]
UW, epoch 6 (PhysioAI) & KiMoRe Overall & 97.80 & 100.00 & 100.00 & 100.00 & 97.33 & $99.03{\pm}1.34$ \\
 & Hard-67 & 14/16 & 13/13 & 16/16 & 15/15 & 6/7 & $94.64{\pm}7.36$; 64/67 \\
 & UI-PRMD Overall & 87.32 & 83.64 & 77.16 & 91.75 & 97.34 & $87.44{\pm}7.69$ \\
\bottomrule
\end{tabular}
\end{table}

\clearpage
\section{Hard-67 Construction and Evaluation}
\label{sec:supp_hard67}

\subsection{Reference probe and training dynamics}

Hard-67 was constructed using plain ST-GCN as the sole reference probe and Dataset Cartography statistics~\cite{swayamdipta_dataset_2020}. PhysioAI, GAP, and the other comparison methods did not contribute to subset membership. The construction input was the ST-GCN training dynamics from the fixed subject-disjoint KiMoRe folds. Because every participant is held out once, each sample belongs to the training partition in the other four folds. A deterministic dynamics loader evaluated every training-partition sample after each of 80 epochs using fixed temporal preprocessing and no random rotation. Consequently, every one of the 400 KiMoRe samples contributed $4\times80=320$ non-augmented true-class probabilities.

Let $p_{i,f,e}$ denote the true-class probability for sample $i$ when it is in training fold $f$ after epoch $e$, and let $\mathcal{T}_i$ contain its 320 observations. The cartography statistics were
\begin{align}
\mu_i &= \frac{1}{320}\sum_{(f,e)\in\mathcal{T}_i}p_{i,f,e},\\
\sigma_i &= \sqrt{\frac{1}{320}\sum_{(f,e)\in\mathcal{T}_i}(p_{i,f,e}-\mu_i)^2}.
\end{align}
Thus confidence $\mu_i$ is the arithmetic mean and variability $\sigma_i$ is the population standard deviation of the same probability trajectory. These quantities use training-partition dynamics only; no held-out prediction, Hard-67 score, or PhysioAI output enters the construction statistics.

\subsection{Frozen selection rule}

The implementation used confidence quantile $q=0.3319$, reported in the main paper at whole-percentile precision as the 33rd percentile. Samples satisfying $\mu_i\leq\tau_{\mu}$ formed the low-confidence group; within this group, samples satisfying $\sigma_i\leq\tau_{\sigma}$ were retained, where $\tau_{\sigma}$ was the median variability. Table~\ref{tab:supp_hard67_construction} gives the thresholds and resulting counts. Retaining low confidence together with low variability targets cases for which the reference model remains consistently uncertain rather than cases whose predictions fluctuate strongly across training.

\begin{table}[h]
\centering
\caption{Hard-67 construction statistics. Thresholds are given at sufficient precision to reproduce the fixed subset.}
\label{tab:supp_hard67_construction}
\setlength{\tabcolsep}{4pt}
\begin{tabular}{lr}
\toprule
Item & Value \\
\midrule
KiMoRe samples & 400 \\
Training partitions per sample & 4 \\
Epochs per training partition & 80 \\
Observations per sample & 320 \\
Implemented confidence quantile $q$ & 0.3319 \\
Confidence threshold $\tau_{\mu}$ & 0.6014788540 \\
Low-confidence samples & 133 \\
Within-group variability threshold $\tau_{\sigma}$ & 0.2713531820 \\
Selected samples & 67 (16.75\%) \\
\bottomrule
\end{tabular}
\end{table}

\subsection{Manifest and distribution}

The complete fixed manifest is supplied as \texttt{assets/hard67/hard67\_manifest.csv}. Each row contains the dataset-derived sample ID, label index, exercise code and name, held-out fold, confidence, variability, and observation count.

Table~\ref{tab:supp_hard67_distribution} summarises the subset using the mapped CKD action names; Es5 corresponds to the KiMoRe exercise \emph{Squatting}. Unequal fold sizes are why the unweighted Hard-67 fold mean and pooled correct/67 accuracy are distinct summaries.

\begin{table}[H]
\centering
\caption{Hard-67 distribution by held-out fold and exercise.}
\label{tab:supp_hard67_distribution}
\begin{tabular}{llr}
\toprule
Dimension & Category & Count \\
\midrule
Held-out fold & F1 & 16 \\
 & F2 & 13 \\
 & F3 & 16 \\
 & F4 & 15 \\
 & F5 & 7 \\
\midrule
Exercise & Es1: Shoulder Flexion & 6 \\
 & Es2: Lateral Trunk Tilt & 5 \\
 & Es3: Trunk Rotation & 14 \\
 & Es4: Pelvis Rotation & 22 \\
 & Es5: Deep Squat & 20 \\
\bottomrule
\end{tabular}
\end{table}

The manifest uses opaque dataset-derived sample IDs to make subset membership reproducible. The dataset participant code is embedded in each sample ID.

\subsection{Held-out evaluation}

Each selected sample is evaluated in its subject's test fold using the checkpoint selected under Section~\ref{sec:supp_protocol}; Hard-67 membership, labels, and scores do not guide checkpoint selection.

Hard-67 aggregation follows Section~\ref{sec:supp_protocol}, and Section~\ref{sec:supp_fold_results} provides the fold-level results for all reported methods and ablations.

Hard-67 should be interpreted as a fixed, ST-GCN-derived stress test for difficult recognition instances. Its results do not define clinical movement difficulty, movement quality, therapeutic correctness, or robustness across all patient populations and execution settings.
\section{Artifact Index and Code Availability}
\label{sec:supp_artifacts}

The submitted ZIP contains this PDF, a top-level \texttt{README.txt}, and the machine-readable assets indexed below.

\begin{table}[h]
\centering
\caption{Relationship between supplementary sections and submitted assets.}
\label{tab:supp_artifact_index}
\scriptsize
\setlength{\tabcolsep}{4pt}
\renewcommand{\arraystretch}{0.96}
\begin{tabularx}{\linewidth}{llX}
\toprule
Section & Asset group & Contents \\
\midrule
S1 & \texttt{assets/ckd/} & Catalogue, prompts, structured entries, candidates, selected sentences, mappings, and selection definitions. \\
S2 & \texttt{assets/protocol/} & Fold manifests, dataset and joint mappings, preprocessing, reproducibility, checkpoint rule, and metrics. \\
S3 & \texttt{assets/methods/} & Common protocol, method recipes, GAP text assets, and exact ablation definitions. \\
S4 & \texttt{assets/results/} & Five-fold values underlying the comparison and ablation tables. \\
S5 & \texttt{assets/hard67/} & Frozen 67-sample manifest, distribution, thresholds, and construction summary. \\
\bottomrule
\end{tabularx}
\end{table}

The CSV and JSON files use UTF-8 encoding. The top-level README gives file descriptions and the aggregation procedure.

Full training and evaluation code will be released publicly upon acceptance.

{\small
\bibliographystyle{ieeenat_fullname}
\bibliography{main}
}